\pdfoutput=1

\documentclass[11pt]{article}

\usepackage[]{acl}

\usepackage{times}
\usepackage{latexsym}

\usepackage[T1]{fontenc}

\usepackage[utf8]{inputenc}

\usepackage{microtype}

\usepackage{hyperref}       
\usepackage{url}            
\usepackage{booktabs}       
\usepackage{amsfonts}       
\usepackage{nicefrac}       
\usepackage{xcolor}         
\usepackage{cite}
\usepackage{amsthm}
\usepackage{amsmath}
\usepackage{amssymb}
\usepackage{bbm}
\usepackage{physics}

\newtheorem{theorem}{Theorem}

\newtheorem{lemma}[theorem]{Lemma}
\newtheorem{proposition}[theorem]{Proposition}

\theoremstyle{definition}
\newtheorem{definition}{Definition}

\newcommand{\squishlist}{
\begin{list}{{{\small{$\bullet$}}}}
{\setlength{\itemsep}{3pt}      \setlength{\parsep}{1pt}
\setlength{\topsep}{1pt}       \setlength{\partopsep}{0pt}
\setlength{\leftmargin}{1em} \setlength{\labelwidth}{1em}
\setlength{\labelsep}{0.5em} } }
\newcommand{\squishend}{  \end{list}  }

\usepackage{array,booktabs,tabularx,multirow,multicol,colortbl,hhline}

\makeatletter
\newcommand{\ccell}[3][]{%
  \kern-\fboxsep
  \if\relax\detokenize{#1}\relax
    \expandafter\@firstoftwo
  \else
    \expandafter\@secondoftwo
  \fi
  {\colorbox{#2}}%
  {\colorbox[#1]{#2}}%
  {#3}\kern-\fboxsep
}
\makeatother

\usepackage{graphicx}
\usepackage{subcaption}
\usepackage[capitalize]{cleveref}

\usepackage{macros}

\title{Assessing Multilingual Fairness in Pre-trained \\ Multimodal Representations}

\author{Jialu Wang \and Yang Liu \and Xin Eric Wang \\
Department of Computer Science and Engineering \\
  University of California, Santa Cruz \\
  \texttt{\{faldict,yangliu,xwang366\}@ucsc.edu}}

\begin{document}
\maketitle

\begin{abstract}
Recently pre-trained multimodal models, such as CLIP~\citep{CLIP}, have shown exceptional capabilities towards connecting images and natural language. The textual representations in English can be desirably transferred to multilingualism and support downstream multimodal tasks for different languages. Nevertheless, the principle of multilingual fairness is rarely scrutinized: do multilingual multimodal models treat languages equally? Are their performances biased towards particular languages? To answer these questions, we view language as the fairness recipient and introduce two new fairness notions, multilingual individual fairness and multilingual group fairness, for pre-trained multimodal models. Multilingual individual fairness requires that text snippets expressing similar semantics in different languages connect similarly to images, while multilingual group fairness requires equalized predictive performance across languages. We characterize the extent to which pre-trained multilingual vision-and-language representations are individually fair across languages. However, extensive experiments demonstrate that multilingual representations do not satisfy group fairness: (1) there is a severe multilingual accuracy disparity issue; (2) the errors exhibit biases across languages conditioning the group of people in the images, including race, gender and age.
\end{abstract}

\section{Introduction}
\label{sec:intro}

Recently pre-trained vision-and-language representations~\citep{Lu2019ViLBERTPT,tan2019lxmert,Su2020VLBERTPO,Li2020UnicoderVLAU,chen2020uniter,Li2020OscarOA,gan2020large,Yu2020ERNIEViLKE,desai2021virtex,CLIP,Cho2021UnifyingVT} have received a surge of attention. Such pre-trained multimodal representations have shown great capabilities of bridging images and natural language on the downstream tasks, including image captioning~\citep{Laina2019TowardsUI}, image retrieval~\citep{Vo2019ComposingTA}, visual QA~\citep{Zhou2020UnifiedVP}, text-to-image generation~\citep{Ramesh2021ZeroShotTG}, etc. While it is commonly recognized that the multimodal representations trained on English corpora can be generalized to multilingualism by cross-lingual alignment~\citep{Lample2019CrosslingualLM,Conneau2020UnsupervisedCR}, recent studies criticize that the multilingual textual representations \emph{do not} learn equally high-quality representations for all the languages~\citep{Wu2020AreAL}, especially for low-resource languages. \citet{hu2020xtreme} emphasize the need for general-purpose representations to seek equal performance across all languages. However, there is still a lack of a nuanced understanding of how multilingual representations fare on vision-and-language benchmarks.

This paper provides a novel perspective for analyzing the principles of multilingual fairness in multimodal representations from two aspects. First, existing frameworks for measuring multilingual biases usually emulate text sources in different languages, which may have ambiguous meanings in varied contexts~\citep{gonzalez-etal-2020-type}. In contrast, we leverage visual grounding as the anchor to bridge text in different languages---text snippets in different languages but with similar semantics should be equitably relevant to the same images. Second, we equate a language as an aggregated group of individuals (e.g., French as a group of French sentences) in the terminology of fairness. As \citet{choudhury2021how} has pointed out, ``\textit{each language has a distinct identity, defined by its vocabulary, syntactic structure, its typological features, amount of available resources, and so on.}'' The notions of fairness, such as individual fairness~\citep{Dwork2012FairnessTA} and group fairness~\citep{Zemel2013LearningFR,Chouldechova2017FairPW,Hardt2016EqualityOO,zhu2022the}, can be naturally adapted by comparing the multimodal model's treatment across languages. 

Therefore, we introduce two fairness notions: \emph{multilingual individual fairness} presumes similar outcomes between similar language expressions grounding on the same images; \emph{multilingual group fairness} postulates that multimodal models should induce similar predictive performance across different languages. These fairness notions are formalized to \emph{compare the multimodal model's treatment of one language versus another} for either the individual target or the aggregated group.

Our contributions are as follows:
\squishlist
    \item We formally define the individual fairness and group fairness notions in the multilingual and multimodal setting (see \cref{sec:individual-fairness-result} and \cref{sec:group-fairness-results}).
    \item We theoretically investigate the extent to which pre-trained multilingual vision-and-language representations are individually fair. However, our negative result demonstrates that individual fairness does not suffice to prevent accuracy disparity at the group level (see \cref{sec:individual-fairness-evaluation}).
    \item Extensive experimental results reveal the accuracy disparity across different languages. Our results also imply that the choice of visual representations affects the group fairness metrics (see \cref{sec:exp-accuracy-disparity}).
    \item We further demonstrate the prevalence of group rate disparity when language is coupled with multi-dimensional groups associated with images, such as race, gender, and age (see \cref{sec:multilingual-group-rate-disparity-result}). 
    Our empirical exploration provides new directions for mitigating biases under the multilingual setting.
\squishend 
\section{Background}\label{sec:background}
\paragraph{Notation.} Throughout the paper, we use the uppercase letter $I$ to denote images and $T$ to denote text. We use the superscript $(L)$ in $T^{(L)}$ to represent the text is in language $L$. When we are jointly using $T^{(L)}$ and $T^{(L')}$ for two languages $L$ and $L'$, we often assume that they share the same semantic meanings. Lowercase letters $\vb*{v}$ and $\vb*{t}$ are used to denote the visual and textual representation vectors encoded by model $M$, respectively. To simplify the presentation, we use $S(\cdot, \cdot)$ to generally represent the similarity between images and text. Specifically, $S(I, T)$ refers to the similarity scores predicted by the model $M$ between the image $I$ and text $T$, while $S(\vb*{v}, \vb*{t})$ refers to the cosine similarity between vectors $\vb*{v}$ and $\vb*{t}$.

\subsection{Multilingual CLIP} 
Our work is established on the multimodal setting. The universal framework for matching images and text~\citep{Mogadala2019TrendsII} is to encode them into representation vectors in a shared representation space, such that the distance between visual and textual vectors can measure the similarity between images and text. Throughout this paper, our analysis mainly focuses on CLIP (Contrastive Language-Image Pre-training~\citealp{CLIP}), a representative pre-trained multimodal representation model that achieves state-of-the-art performances on zero-shot transfer tasks.

CLIP is a multimodal model trained on large-scale images with natural language supervision collected from the internet. It comprises an image encoder and a text encoder that can embed images and text into visual and textual representation vectors. One desirable property is that the CLIP model takes the cosine similarity between image and text features to measure the log-odds of the corresponding image-text pairs, and is trained to maximize their similarity by a contrastive learning objective. In light of this capability, CLIP can predict the similarity, denoted by $S(I, T)$, between arbitrary images $I$ and natural language text snippets $T$. 

In order to adapt the flexible CLIP model to multilingualism, Multilingual CLIP~\citep{Multilingual-CLIP} uses a pre-trained multilingual language model, such as M-BERT~\citep{Devlin2019BERTPO}, to take over the original text encoder in English, and fine-tune the textual representation vectors by cross-lingual alignment~\citep{Lample2019CrosslingualLM,Conneau2020UnsupervisedCR}. In this setting, we use $S(I, T^{(L)})$ to represent the similarity between image $I$ and text $T^{(L)}$ in language $L$. Though the empirical evaluations in this paper mainly focus on Multilingual CLIP, the experimental approaches we adopt to arrive at the observations can be generalized to other pre-trained multilingual vision-and-language representations. 

\subsection{Fairness Notions} 
The multilingual fairness notions developed in this work is inspired by multiple fairness definitions~\citep{narayanan2018translation,Dwork2012FairnessTA} in the algorithmic fairness literature. 
We will briefly introduce these fairness notions in fair decision making and instantiate them in the domain of multilingual vision-and-language learning later.

Individual fairness, initiated by \citet{Dwork2012FairnessTA}, requires that individuals who are similar with respect to a task-specific similarity metric have similar decision outcomes.

Group fairness definitions seek to provide fairness guarantees based on group-level statistical constraints, in the sense that they are evaluated and enforced without reference to similarity measures. In the fairness literature, group fairness is commonly framed in terms of \emph{protected groups} $G$, such as race, gender, and age. For instance, demographic parity~\citep{Zafar2017FairnessCM} requires that the outcomes are independent of the group membership, and equalized odds~\citep{Hardt2016EqualityOO} essentially requires equal true positive and false positive rates between different groups. 

Principally, these fairness criteria are formulated by comparing the treatment of one individual or one group versus another. Our work will instantiate the standard fairness notions by viewing language as the recipient --- we compare \emph{how the treatment of one language differs from another}.

\subsection{Fairness in NLP} 
Many recent works~\citep{choudhury2021how,hu2020xtreme,Libovick2019HowLI,gonzalez-etal-2020-type,Ross2021MeasuringSB} scrutinize the ethical issues raised in multilingual settings, albeit with varying degrees of success. For instance, \citet{zhao-etal-2020-gender} quantifies the presence of representational biases in multilingual word embeddings by calculating the distance between targets corresponding to different sensitive attributes. \citet{huang-etal-2020-multilingual} evaluate group fairness violations among demographic groups on the task of hate speech detection, but do not explicitly regard language as unique group membership. \citet{Burns2020LearningTS} studies the performance degradation when multimodal models are trained to support additional languages, and tries to address the multilingual accuracy disparity on the task of image-sentence retrieval. Our work complements the fairness discourse in multilingual NLP to the extent that we provide a novel perspective of studying multilingual fairness by viewing language as the recipient of fairness notions.

Our work is also closely relevant to prior studies on biases in vision-and-language tasks, including visual semantic role labeling~\citep{Zhao2017MenAL}, image captioning~\citep{Burns2018WomenAS,10.1145/3442381.3449950}, and image search~\citep{Wang2021MitigateGenderBiasInImageSearch}. Notably, \citet{Srinivasan2021WorstOB} investigates the gender bias associated with entities for pre-trained representations. Compared to these works, we focus on generic fairness measures for multimodal models and use visual grounding to bridge different languages.
\section{Multilingual Individual Fairness}
\label{sec:individual-fairness-result}
For an ideal multilingual vision-and-language model, text descriptions in different languages referring to similar semantic meanings should be equally similar or dissimilar to the same grounding images. We note that there are no language expressions that are perfectly identical to each other in real-world scenarios due to linguistic features. Nevertheless, at least in a normal vision-and-language task, multilingual models are desired to impose equal treatment to different languages. For instance, ``this is a cat'' (in English) and ``das ist eine Katze'' (in German) should be similarly related to an image of a cat in image-text retrieval. This intuition aligns with individual fairness in a multilingual manner. In this section, we investigate to what degree multilingual representations are individually fair.

Individual fairness requires that similar people should be treated similarly \citep{Dwork2012FairnessTA}. In our multilingual setting, we require that the text snippets expressing similar semantics in different languages should be similarly related to the same images. Taking the Euclidean distance function to measure the distance between text features, we can define $\alpha$-multilingual individual fairness by: 

\begin{definition}[Multilingual Individual Fairness]\label{def:multilingual-individual-fairness}
Given a set of image-text pairs $\{(I, T)\}$, a multimodal model $M$ satisfies $\alpha$-multilingual individual fairness if for all $(I, T)$, for languages $L$ and $L'$:
\[
    |S(I, T^{(L)}) - S(I, T^{(L')})| \leq \alpha \|\vb*{t}^{(L)} - \vb*{t}^{(L')}\|
\]
where $\vb*{t}^{(L)}$ is the textual representation vector yielded by $M$ in language $L$.
\end{definition}
Here, $\alpha$ is a parameter to control the ratio of similarity gap to the text feature vectors' distance, and smaller $\alpha$ indicates the model is individually fairer. Note that the similarity gap is at most 2, because the range of cosine similarity is $[0, 1]$. In general settings, $S(I, T)$ is measured by the cosine similarity between the encoded visual vector $\vb*{v}$ and textual vector $\vb*{t}$.
\begin{lemma}\label{lem:individual-fairness}
Denote $\mathcal{O}_{\rho} (\vb*{t}) = \{ \vb*{x} \mid  \|\vb*{x} - \vb*{t} \| \leq \rho \}$ to be a closed ball of radius $\rho > 0$ and center $\vb*{t}$. Then for any visual representation vector $\vb*{v}$,
\begin{multline}
\sup_{\small \substack{\vb*{t}^{(L')} \in \mathcal{O}_{\rho}(\vb*{t}^{(L)}) \\ 0 \leq \rho < \|\vb*{t}^{(L)}\|}} |S(\vb*{v}, \vb*{t}^{(L')}) - S(\vb*{v}, \vb*{t}^{(L)})|  \\ \leq \sqrt{2 (1 - \sqrt{1 - (\frac{\rho}{\|\vb*{t}^{(L)}\|})^2})}
\end{multline}
where $S(\cdot, \cdot)$ denotes the cosine similarity, $\vb*{t}^{(L)}$ and $\vb*{t}^{(L')}$ are textual representation vectors for languages $L$ and $L'$, respectively. 
\end{lemma}
We defer the proof to \cref{proof:individual-fairness}. Lemma~\ref{lem:individual-fairness} implies that when the distance between multilingual textual representation vectors is bounded, the similarity with images can be bounded in terms of their distance. It is worth noting that the bounds are independent of the visual representation vectors. Nevertheless, the form of upper bound in \cref{lem:individual-fairness} is a bit sophisticated, and can be simplified when $\rho \ll \|\vb*{t}^{(L)}\|$. 
\begin{theorem}\label{thm:individual-fairness-approximation}
When $\|\vb*{t}^{(L')} - \vb*{t}^{(L)}\| \ll \|\vb*{t}^{(L)}\|$, 
\[
|S(\vb*{v}, \vb*{t}^{(L')}) - S(\vb*{v}, \vb*{t}^{(L)})| \lessapprox \frac{\|\vb*{t}^{(L')} - \vb*{t}^{(L)}\|}{\|\vb*{t}^{(L)}\|}.
\]
\end{theorem}
Theorem~\ref{thm:individual-fairness-approximation} is a direct application of Lemma~\ref{lem:individual-fairness} when the distance between multilingual vectors is small enough, and extends in many natural cases to approximate the multilingual individual fairness with $\alpha \approx \frac{1}{\|\vb*{t}^{(L)}\|}$. The proof can be found in \cref{proof:individual-fairness-approximation}. Theorem~\ref{thm:individual-fairness-approximation} implicates to what degree the multimodal model satisfies individual fairness when text snippets are well aligned between different languages.
\section{Multilingual Group Fairness}\label{sec:group-fairness-results}
Distinct from individual fairness, multilingual group fairness appeals to the idea that multimodal models should achieve equivalent predictive performance across different languages. From the perspective of representations, it is hard to carry out this demand without well-defined tasks and metrics. Hence it is natural to ask how to define group fairness in this scenario properly? In this section, we shall answer this question by equating language as a unique dimension of group membership relating to the text modality. We formulate the criteria by equalizing the accuracy rates over different languages. We also observe that images are often connected to people in protected or unprotected groups. Given the image-text pairs, we consider the accuracy disparity across different languages conditioning the subgroup of images. 

\subsection{Equality of Accuracy across Languages}
Given a dataset $\mathcal{D}$ consisting of ground-truth image-text pairs $\{(I_i, T_i)\}$ and each text can be in different languages. The goal of a multimodal model $M$ is to predict the similarity $S(I_i, T_j)$ for any image $I_i$ and text $T_j$. Then the model matches $\hat{T}_i$ for images $I_{i}$ by selecting the text with highest similarity scores, i.e., $\hat{T}_i = \argmax_j S(I_i, T_j)$.
\begin{align}
    \acc (M) & = \frac{1}{|\mathcal{D}|} \sum_{\mathcal{D}} \indicator[\hat{T}_i = T_i] 
\end{align}
We use the superscript $(L)$ to indicate the accuracy $\acc^{(L)}$ is evaluated in language $L$. Next, we take language as group membership and define multilingual accuracy parity by equalizing accuracy across languages.

\begin{definition}[multilingual accuracy parity]\label{def:accuracy-parity} A multimodal model $M$ satisfies \textit{multilingual accuracy parity} if $\acc^{(L)}(M) = \acc^{(L')}(M)$ for all languages $L$, $L'$.
\end{definition}
In practice, it is impossible to achieve accuracy parity for all languages. Following~\citep{hu2020xtreme}, we use
\begin{equation}\label{eq:gap-definition}
    \gap_M (L, L') = |\acc^{(L)}(M) - \acc^{(L')}(M)|
\end{equation}
to represent the cross-lingual gap for model $M$. 

\subsection{When Language Meets Groups in Images}
The above discussion on group fairness considers language as the sole group membership. In the real-world image and text applications, the people portrayed in the images are often associated with protected groups. For instance, the face attribute dataset~\citep{CelebA} contains sensitive attributes, such as race, age and gender. Let $G$ denote the group membership of images and $\mathcal{D}_a$ denote the subset of data examples $\mathcal{D}$ given $G = a$. The accuracy of a multimodal model evaluated on the images of subgroup $a$ is defined as
\begin{align}
    \acc_a (M) & = \frac{1}{|\mathcal{D}_a|} \sum_{\mathcal{D}_a} \indicator[\hat{T}_i = T_i]
\end{align}
When language is connected to images of different groups, we can define accuracy disparity between group $a$ and group $b$ with respect to model $M$ within language $L$ as
\begin{equation}\label{eq:disp-definition}
    \disp_M^{(L)}(a, b) = |\acc^{(L)}_a(M) - \acc^{(L)}_b(M)|
\end{equation}
$\disp$ represents the group rate gap in a single language. Mirroring \textit{multilingual accuracy parity}, we can define the \textit{multilingual group rate parity} as below.
\begin{definition}[multilingual group rate parity]\label{def:group-rate-parity} A multi-modal model $M$ satisfies multilingual group rate parity if $\disp_M^{(L)}(a, b) = \disp_M^{(L')}(a, b)$ with respect to groups $a, b$ associated with images for all languages. 
\end{definition}

\cref{def:accuracy-parity} and \cref{def:group-rate-parity} evaluate the fairness of multilingual representations from diverse aspects. More broadly, we may be interested in the accuracy gap between different combinations of languages and groups. A common case is that there are only two protected groups (e.g. female and male, young and old). Let $p_a = \frac{|\mathcal{D}_a|}{|\mathcal{D}|}$ and $p_b = \frac{|\mathcal{D}_b|}{|\mathcal{D}|}$ represent the population proportions of group $a$ and group $b$ respectively, satisfying $p_a + p_b = 1$. Then we can decompose the cross-lingual cross-group accuracy disparity as below:

\begin{proposition}\label{prop: group-fairness} 
When there are only two protected groups $a$ and $b$, the following inequality holds for any two languages $L$ and $L'$ 
\begin{multline}
|\acc^{(L)}_{a} - \acc^{(L')}_{b}| \leq \gap(L, L') \\ +  p_b \cdot \disp^{(L)}(a, b) + p_a \cdot \disp^{(L')}(a, b)
\end{multline}
\end{proposition}

The proof can be found in \cref{proof:group-fairness}. Proposition~\ref{prop: group-fairness} guarantees that the accuracy disparity between any combinations of languages and protected groups can be upper bounded by a variety of factors, and implicates that we only need to focus on cross-lingual gap and group rate gap measures to assess multilingual group fairness. In what follows, we will take a closer look at how the multilingual CLIP model performs with compositions of languages and protected groups under these fairness criteria.
\section{Evaluations}
In this section, we work with the pre-trained multilingual CLIP \citep{Multilingual-CLIP} model to study multilingual fairness. We validate the extent to which the model is individually fair across different languages in \cref{sec:individual-fairness-evaluation}. We characterize the prevalence of multilingual group unfairness on human faces in \cref{sec:exp-accuracy-disparity} and \cref{sec:multilingual-group-rate-disparity-result}. These empirical evaluations shed light on potential directions for mitigating unfairness in multilingual multimodal representations.

\subsection{Multilingual Individual Fairness}\label{sec:individual-fairness-evaluation}
The theoretical analysis on multilingual individual fairness posed in \cref{sec:individual-fairness-result} implies that the ratio of similarity difference to their text feature distance can be bounded by the reciprocal of the length of text feature vectors. To verify the implication, we conduct experiments on the Multi30K dataset~\citep{elliott-etal-2016-multi30k}.

\paragraph{Dataset.} The Multi30K dataset~\citep{elliott-etal-2016-multi30k} contains 31,014 Flickr30K~\citep{young-etal-2014-image} images and composes the \textit{translation} and the \textit{independent} portions of English-German caption pairs. The German translations were collected from professional English-German translators by translating the English captions without seeing the images, one per image. The independent portion was independently annotated by German crowdworkers after seeing the images instead of English captions, five per image. Hence, the translated captions are strongly aligned in both languages, while the independent descriptions may have distinct context. We use 1,000 test images for our evaluation. For the independent portion, we select the first English caption and the first German caption of the five to pair with the image for a fair comparison.

\begin{figure*}[!htb]
    \centering
    \begin{subfigure}[t]{0.34\linewidth}
        \includegraphics[width=\linewidth]{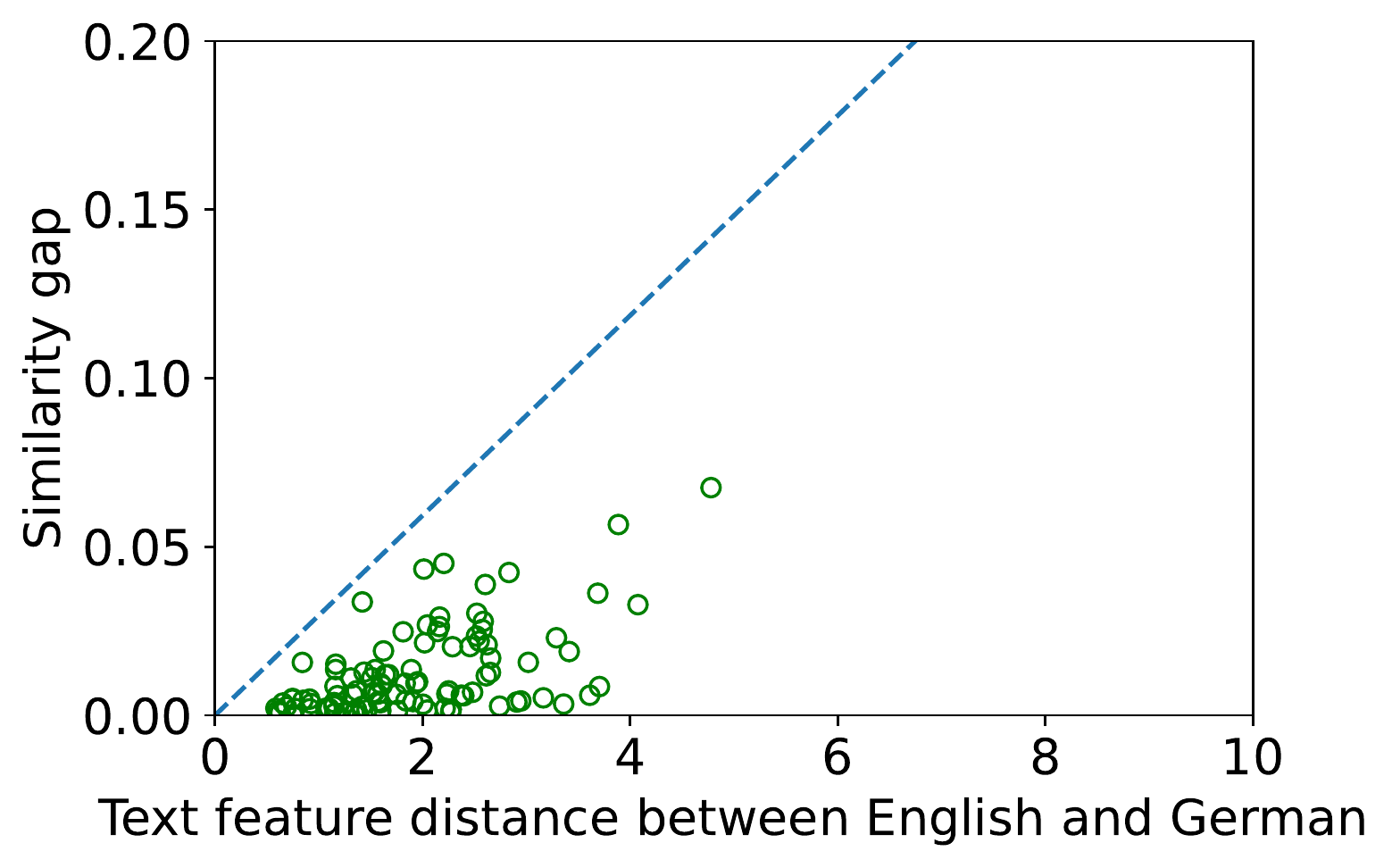}    \caption{Translation}\label{fig:individual-fairness-translation}
    \end{subfigure}
    \hfill
    \begin{subfigure}[t]{0.34\linewidth}
        \includegraphics[width=\linewidth]{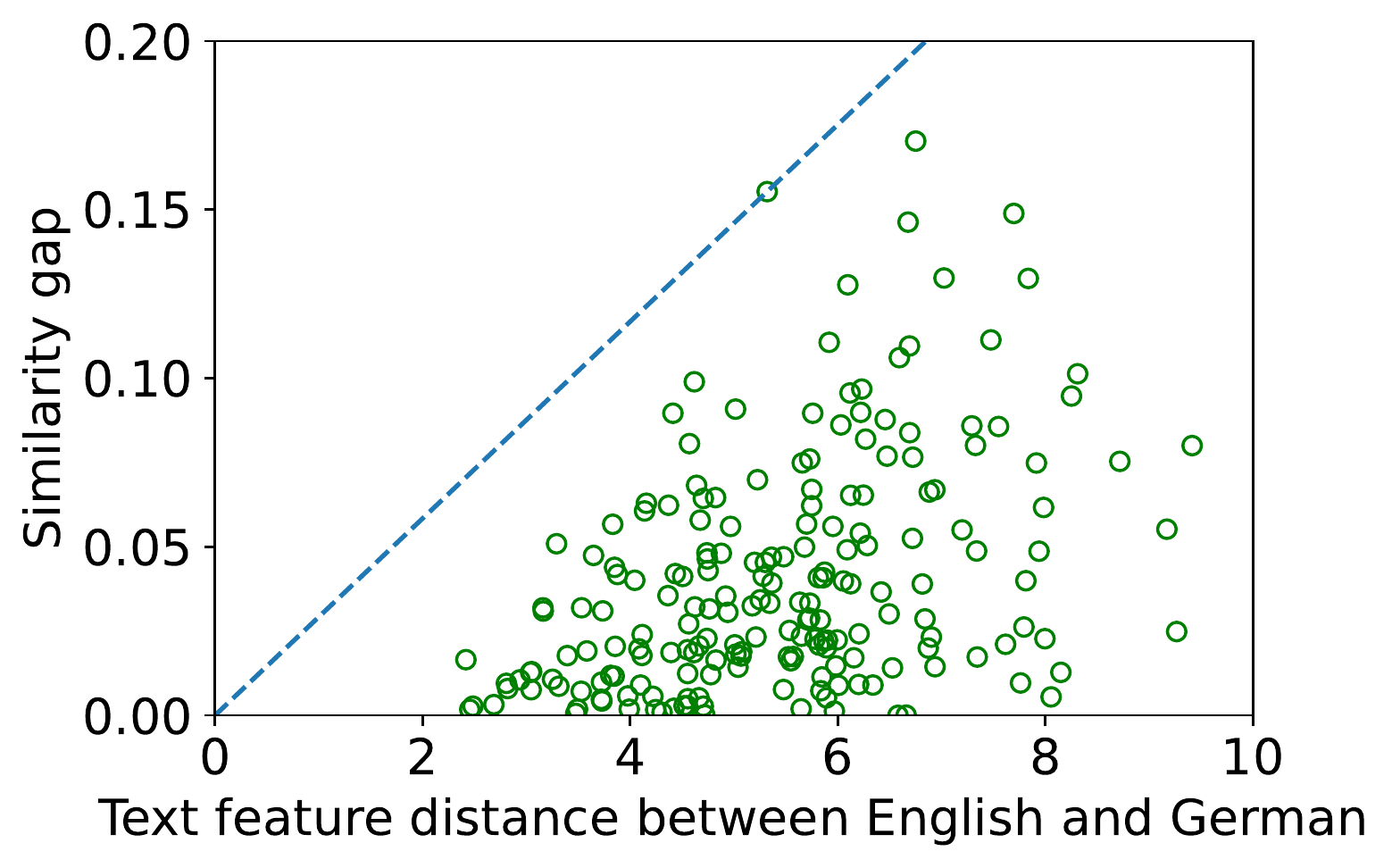}    \caption{Independent}\label{fig:individual-fairness-comparable}
    \end{subfigure}
    \hfill
    \begin{subfigure}[t]{0.3\linewidth}
    \vspace{-1in}
    \resizebox{\linewidth}{!}{
        \begin{tabular}{c | c c}
             & translation & independent  \\
            \hline
            en & 50.4 & 59.2 \\
            de & 45.6 & 36.3 \\
            \hline
            $\gap$ & 4.8 & 22.9 \\
        \end{tabular}}
        \vspace{0.3in}
        \caption{Accuracy disparity}\label{fig:multi30k-accuracy-disparity}
    \end{subfigure}
    \caption{\textbf{We empirically examine how does the multilingual CLIP fare on the translation and the independent portions.} Fig. (\subref{fig:individual-fairness-translation}) and (\subref{fig:individual-fairness-comparable}): the $x$-axis represents the distance between English and German captions, the y-axis represents the gap between their corresponding similarity scores, and the slope of blue dashed lines represents the empirical $\alpha$ for multilingual individual fairness. Fig. (\subref{fig:multi30k-accuracy-disparity}): we evaluate the accuracy for image-text matching, and find out that the independent portion incurs huge accuracy disparity compared with the translation portion.
    }
    \label{fig:individual-fairness}
\end{figure*}
\paragraph{Results.} We embed each English-German caption pair into textual representation vectors and the corresponding image into visual representation vectors. We compute the Euclidean distance between English-German text features, as well as the cosine similarity with respect to the image features. We plot their cross-lingual gap on the translation and the independent portions in \cref{fig:individual-fairness-translation} and \cref{fig:individual-fairness-comparable}, respectively. For both portions, the blue dashed lines represent the empirical upper bounds of the ratio between similarity gap and text feature distance.

Unsurprisingly, we find out that the English-German captions are more closely aligned on the translation portion (the average textual feature distance is 1.86) than the independent portion (average distance is  5.69). The similarity gaps regarding the translation portion are below 0.06 in general, and those regarding the independent portion are above 0.10 for many instances. The reason is apparent: translated captions have more similar semantics owning to the professional text-to-text translations, while independent captions have more diverse expressions of the same images, even if they might refer to the same content.

On the other hand, we observe that the slopes of blue dashed lines in \cref{fig:individual-fairness-translation} and \cref{fig:individual-fairness-comparable} are approximate to each other, i.e., the empirical $\alpha$ for both portions are similar. This fact implies that the multilingual CLIP model evaluated on two different text corpora share a similar level of individual fairness, even though the cross-lingual similarity gaps are quite different. We also note that the empirical upper bound of $\alpha$ are much smaller than the theoretical upper bound $\frac{1}{\|\vb*{t}^{(L)}\|}$ in \cref{thm:individual-fairness-approximation}.

Although we have verified that multilingual multimodal representations satisfy similar individual fairness, we demonstrate that they violate group fairness by evaluating their image-text matching accuracy. As shown in \cref{fig:multi30k-accuracy-disparity}, English captions dominate the Top-1 image-text matching accuracy over German captions, with $4.8\%$ higher on the translation portion and $22.9\%$ higher on the independent portion. This observation delivers an important message for researchers who are interested in learning fair representations~\citep{Ruoss2020LearningCI}: individual fairness \emph{does not} flatly prevent accuracy disparity among different languages~\citep{10.1145/3351095.3372864}.

\begin{figure*}[!t]
    \centering
    \begin{subfigure}[b]{0.3\textwidth}
        \includegraphics[width=\linewidth]{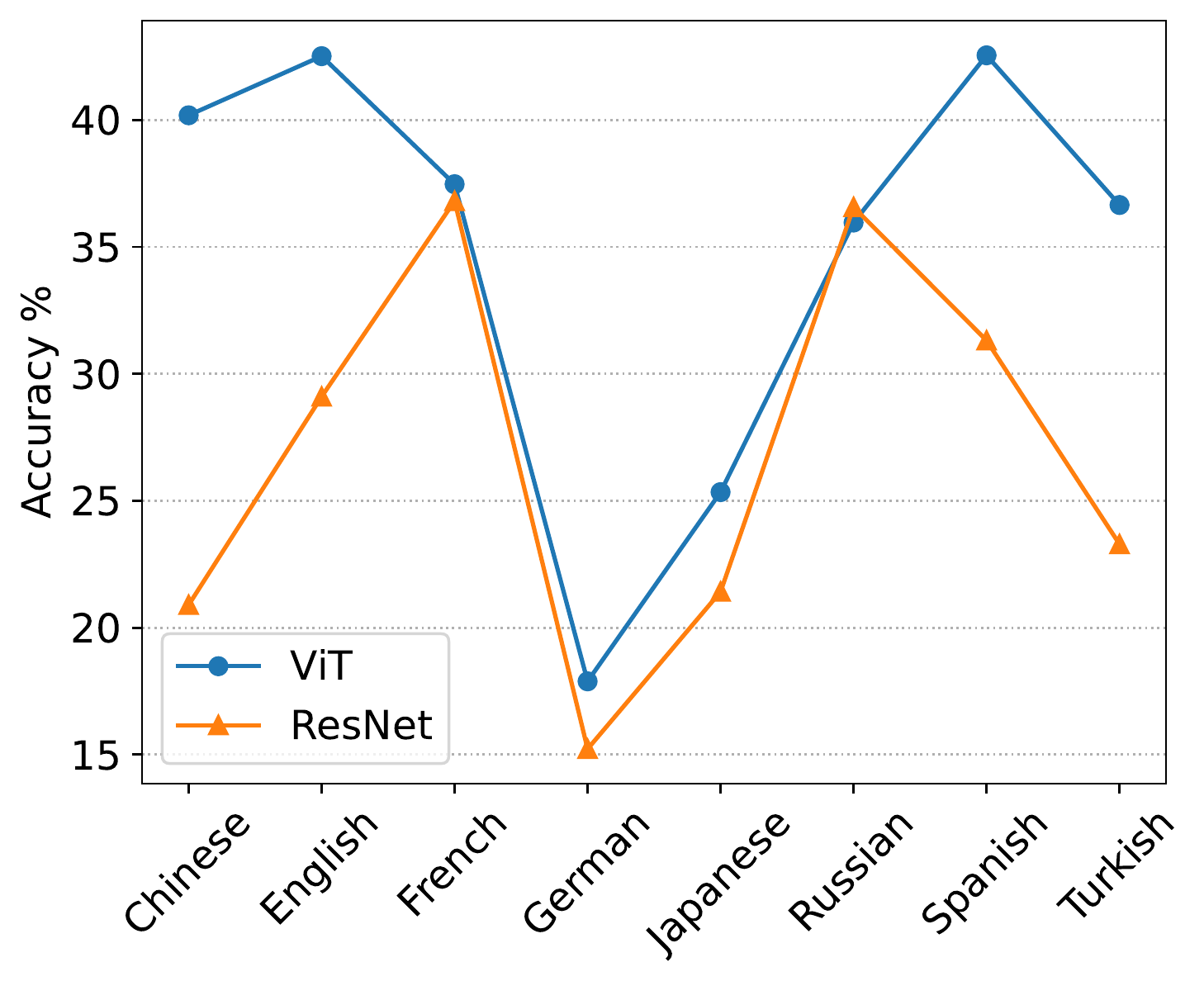}
        \caption{Race classification}
    \end{subfigure}
    \hfill
    \begin{subfigure}[b]{0.3\textwidth}
        \includegraphics[width=\linewidth]{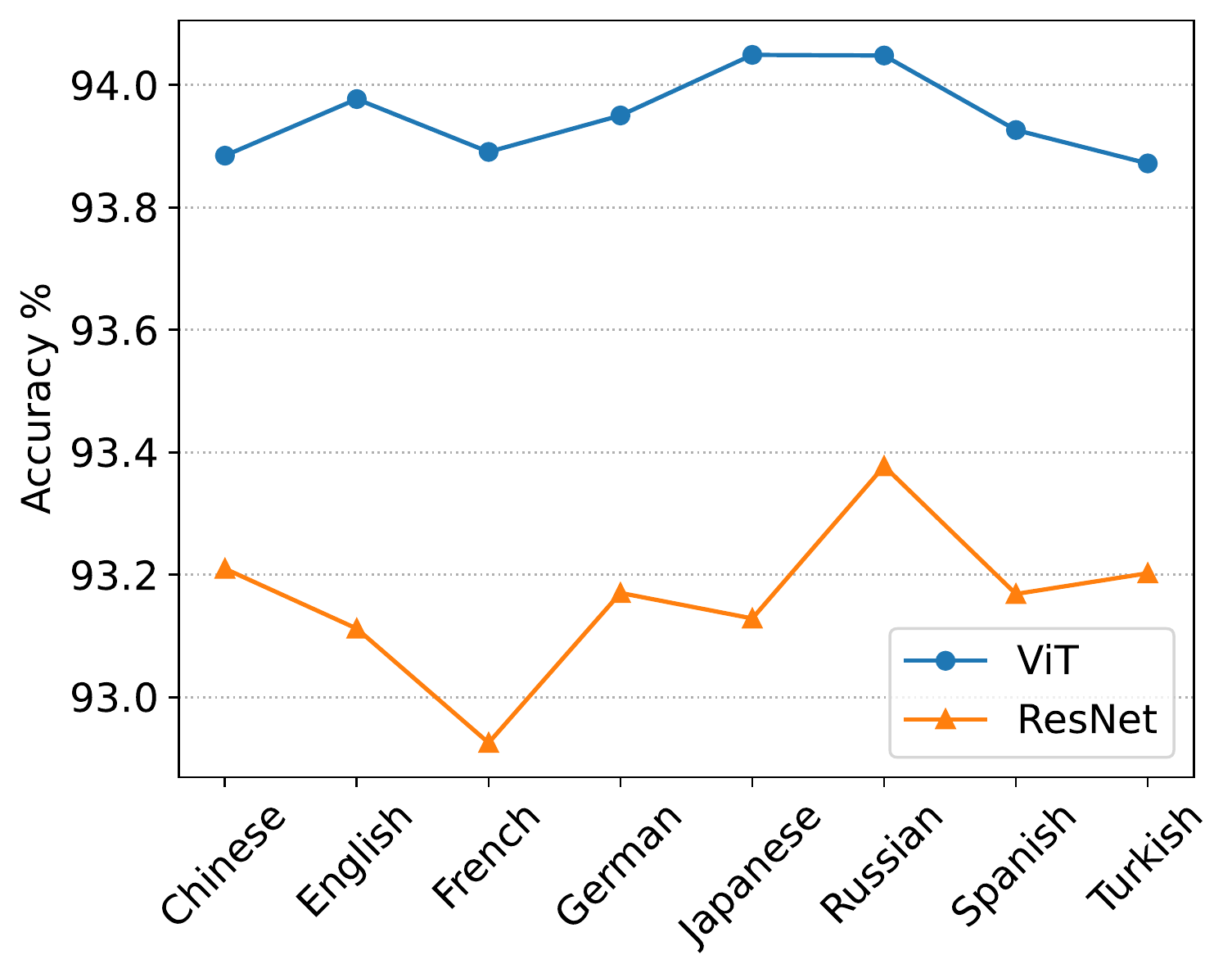}
        \caption{Gender classification}
    \end{subfigure}
    \hfill
    \begin{subfigure}[b]{0.3\textwidth}
        \includegraphics[width=\linewidth]{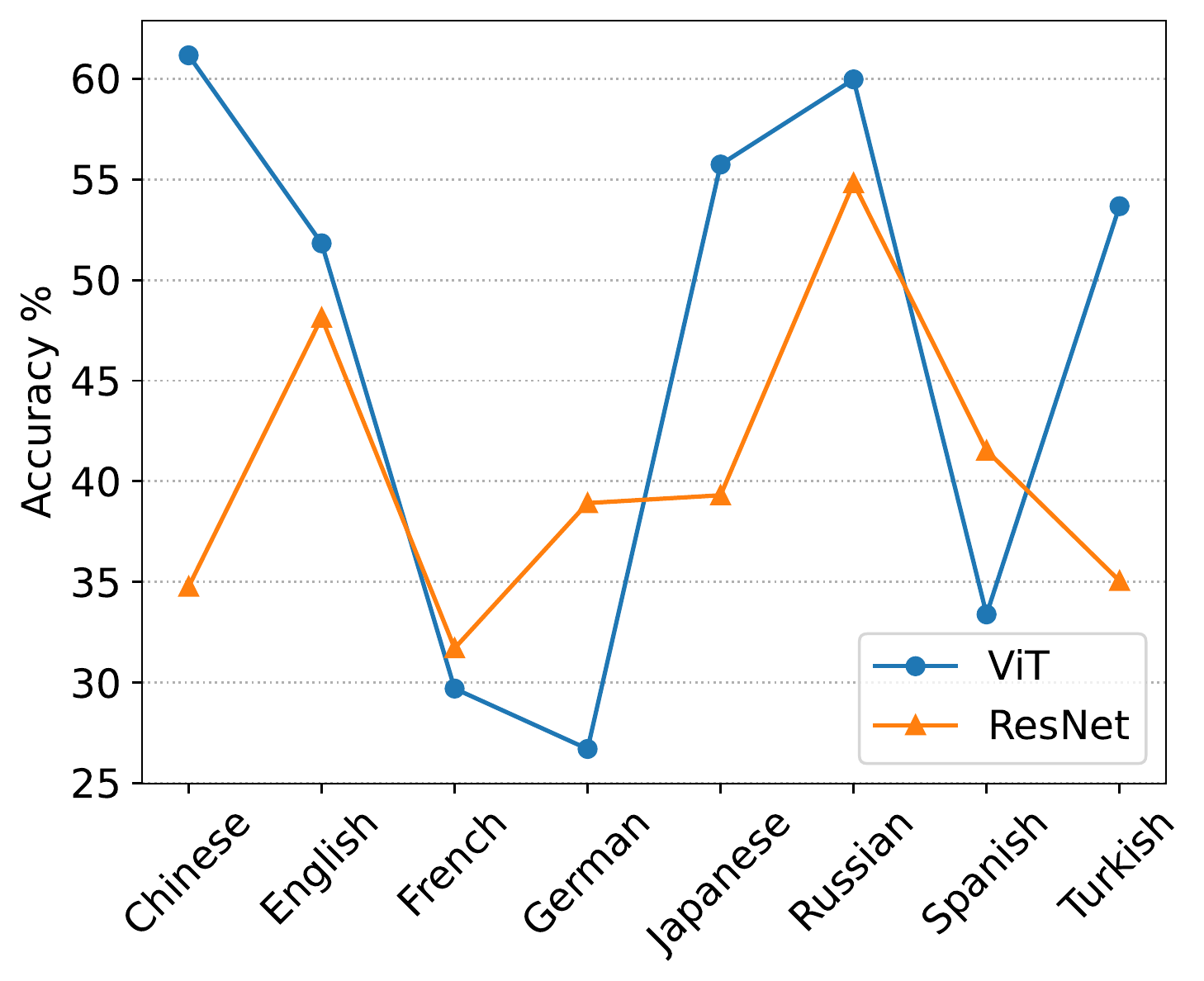}
        \caption{Age classification}
    \end{subfigure}
    \caption{\textbf{Race, gender, and age classification accuracy across different languages.} The languages are in alphabetical order. Two different vision encoders for encoding image features are evaluated, including Vision Transformer (ViT)~\citep{ViT} and ResNet-50 (ResNet)~\citep{ResNet}.
    }
    \label{fig:Race-Prediction-ViT-ResNet}
    \vspace{-0.2in}
\end{figure*}

\subsection{Multilingual Accuracy Disparity}\label{sec:exp-accuracy-disparity}
\paragraph{Dataset.} FairFace~\citep{Krkkinen2019FairFaceFA} is a face attribute dataset for the balanced race, gender, and age groups. It categorizes gender into two groups, including female and male, and race into seven groups, including White, Black, Indian, East Asian, Southeast Asian, Middle Eastern, and Latino. For ages, we categorize the raw labels into five groups: infants (0–2), children and adolescents (2–19), adults (20–49), middle age adults (50–69), and seniors (more than 70). We follow their original data split and select the validation set consisting of 10,940 face images for evaluation.

\paragraph{Languages.} We analyze the multilingual group fairness for 8 languages: Chinese (zh), English (en), French (fr), German (de), Japanese (ja), Russian (ru), Spanish (es), and Turkish (tr). We select English as the pivot language and write natural language prompts in English. Then we translate them into other languages: we first use Google Translate and then recruit native speakers to rate the prompts and fix any potential errors on Amazon Mechanical Turk (see Appendix \ref{sec:AMT} for more details). 
The rationale for only using English as the pivot language is that the multilingual CLIP~\citep{Multilingual-CLIP} selects English as the pivot language for aligning multilingual text embeddings.

\paragraph{Text Prompts.} Following \citet{CLIP}, we construct the text prompt by the template ``A photo of a \texttt{\{label\}} person''. Concretely, for gender classification, we construct the text prompt ``A photo of a woman'' when the gender attribute is female, and construct ``A photo of a man'' otherwise. For race classification, we construct the text prompt by ``A photo of a(n) \{race\} person''. Note that Indian actually refers to South Asian ethnic groups in the Fairface race taxonomy~\citep{Krkkinen2019FairFaceFA} but it can refer to Native Americans as well. To avoid ambiguity, we replace ``Indian'' by ``South Eastern'' to construct the prompts. For age classification, we notice that the age attributes in Fairface dataset are numeric values and use the template ``A photo of a person aged \texttt{\{age\}} years'' to construct text prompts.

\paragraph{Results.} We probe the multilingual accuracy disparity for race classification, gender classification, and age classification, as shown in \cref{fig:Race-Prediction-ViT-ResNet}. We use two different pre-trained image encoders for extracting visual representation vectors, including Vision Transformer~\citep{ViT} and ResNet-50~\citep{ResNet}. We observe that:
\squishlist
\item \textbf{Cross-lingual gap varies across different protected groups.} The predictive accuracy for gender classification is consistently higher than $90\%$ across all the languages. In contrast, the multimodal model has relatively poor performance and more considerable variance for race and age classification. Furthermore, race classification yields $24.66\%$ accuracy disparity and age classification yields $34.47\%$ accuracy disparity for Vision Transformer. This implies that the huge disparity may result from the poor predictive performance of the model.
\item \textbf{Visual representations affect accuracy disparity.} For race classification, Vision Transformer features generally achieve higher accuracy across all languages than ResNet-50 ($34.82\%$ \textit{vs.} $26.83\%$ on average) except for Russian. The standard deviation of Vision Transformer is higher than ResNet-50 ($8.18\%$ \textit{vs.} $7.34\%$). The maximal accuracy gap for Vision Transformer is $30.40\%$ between German and Spanish, while the maximal accuracy gap for ResNet-50 is $23.12\%$ between German and French. For gender classification, Vision Transformer dominantly achieves higher accuracy and incurs less accuracy gap. For age classification, the accuracy is moderately low for all languages. However, Vision Transformer has $63.1\%$ accuracy in Chinese while only $25.8\%$ accuracy in German, exaggerating the accuracy gap between languages.
\squishend
In \cref{tab:gender-classification}, we present the complete results of \cref{fig:gender_gap_across_race_languages} by compositions of gender and race groups across different languages.

\subsection{Multilingual Group Rate Disparity}\label{sec:multilingual-group-rate-disparity-result}
\begin{figure*}[!ht]
    \centering
    \begin{subfigure}[b]{0.24\textwidth}
        \includegraphics[width=\linewidth]{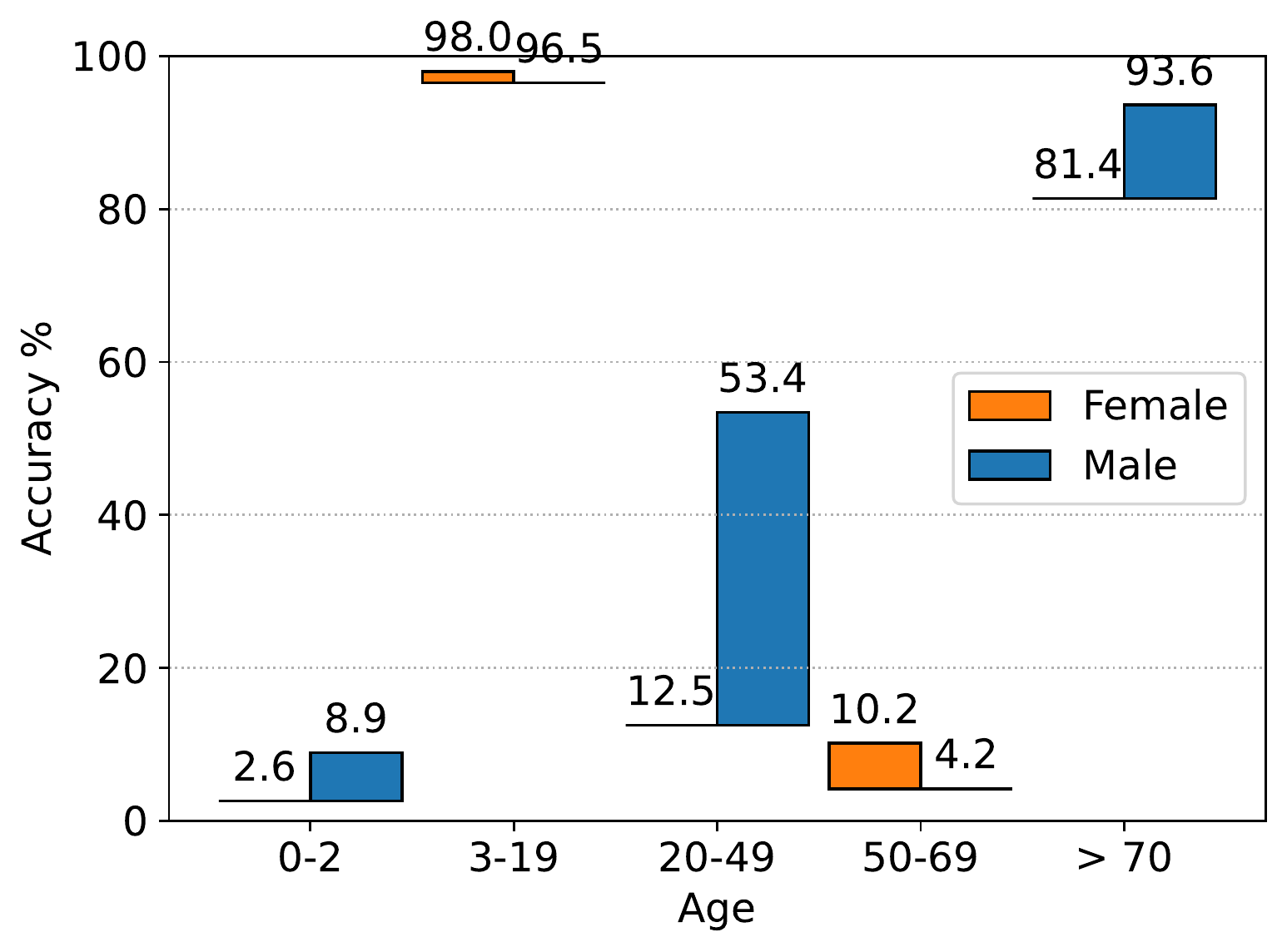}
        \caption{English}
    \end{subfigure}
    \hfill
    \begin{subfigure}[b]{0.24\textwidth}
        \includegraphics[width=\linewidth]{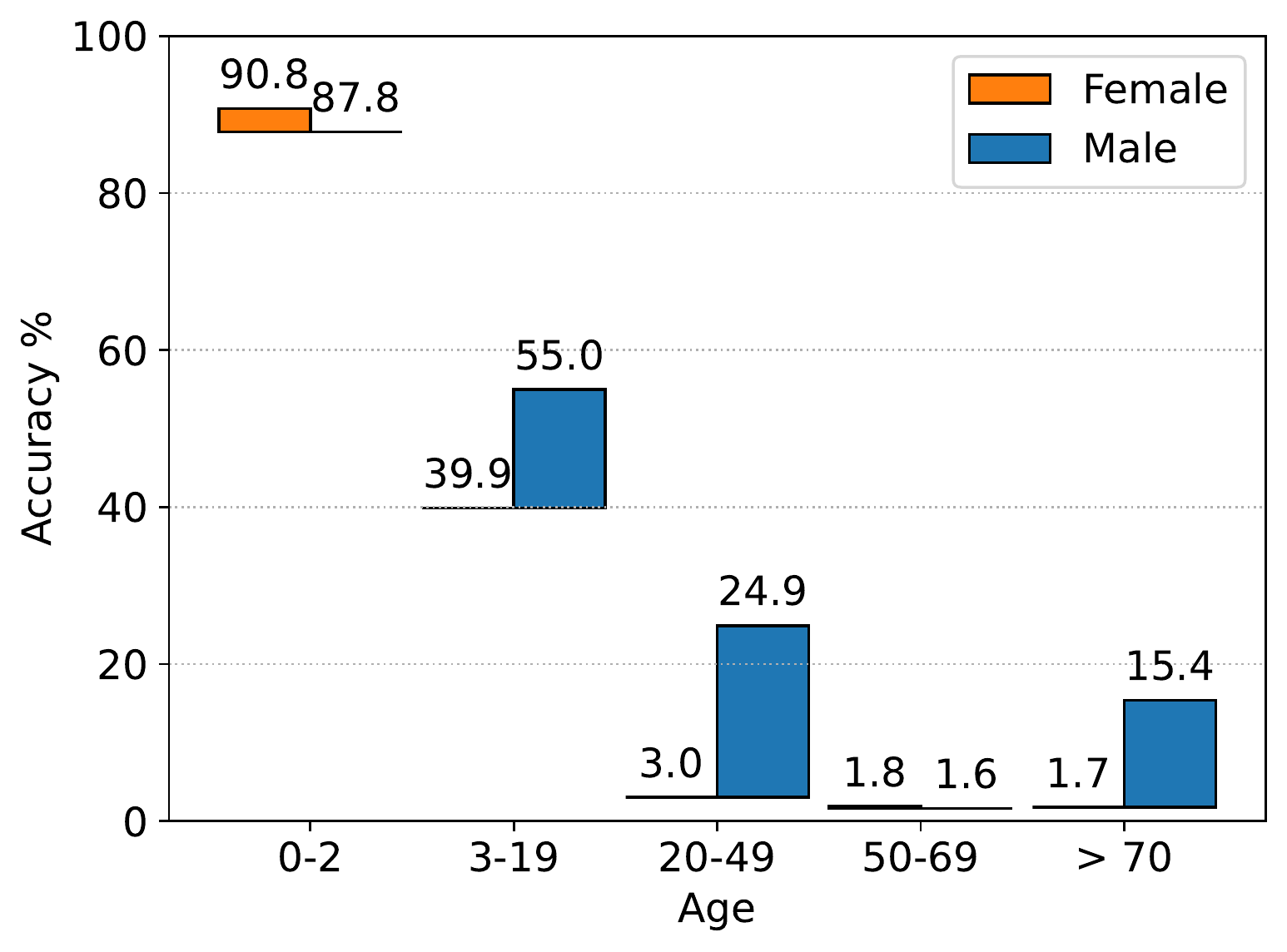}
        \caption{German}
    \end{subfigure}
    \hfill
    \begin{subfigure}[b]{0.24\textwidth}
        \includegraphics[width=\linewidth]{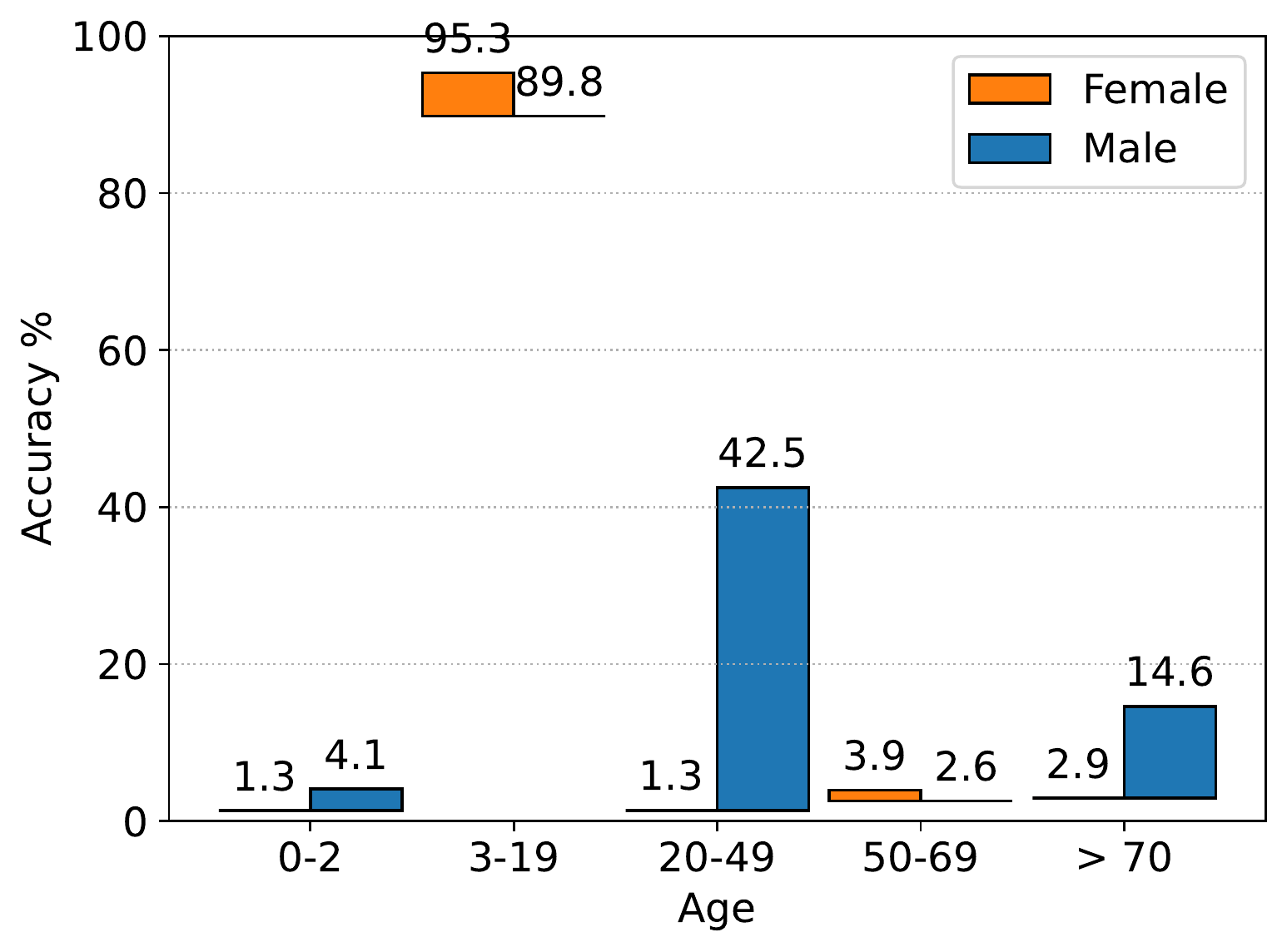}
        \caption{French}
    \end{subfigure}
    \hfill
    \begin{subfigure}[b]{0.24\textwidth}
        \includegraphics[width=\linewidth]{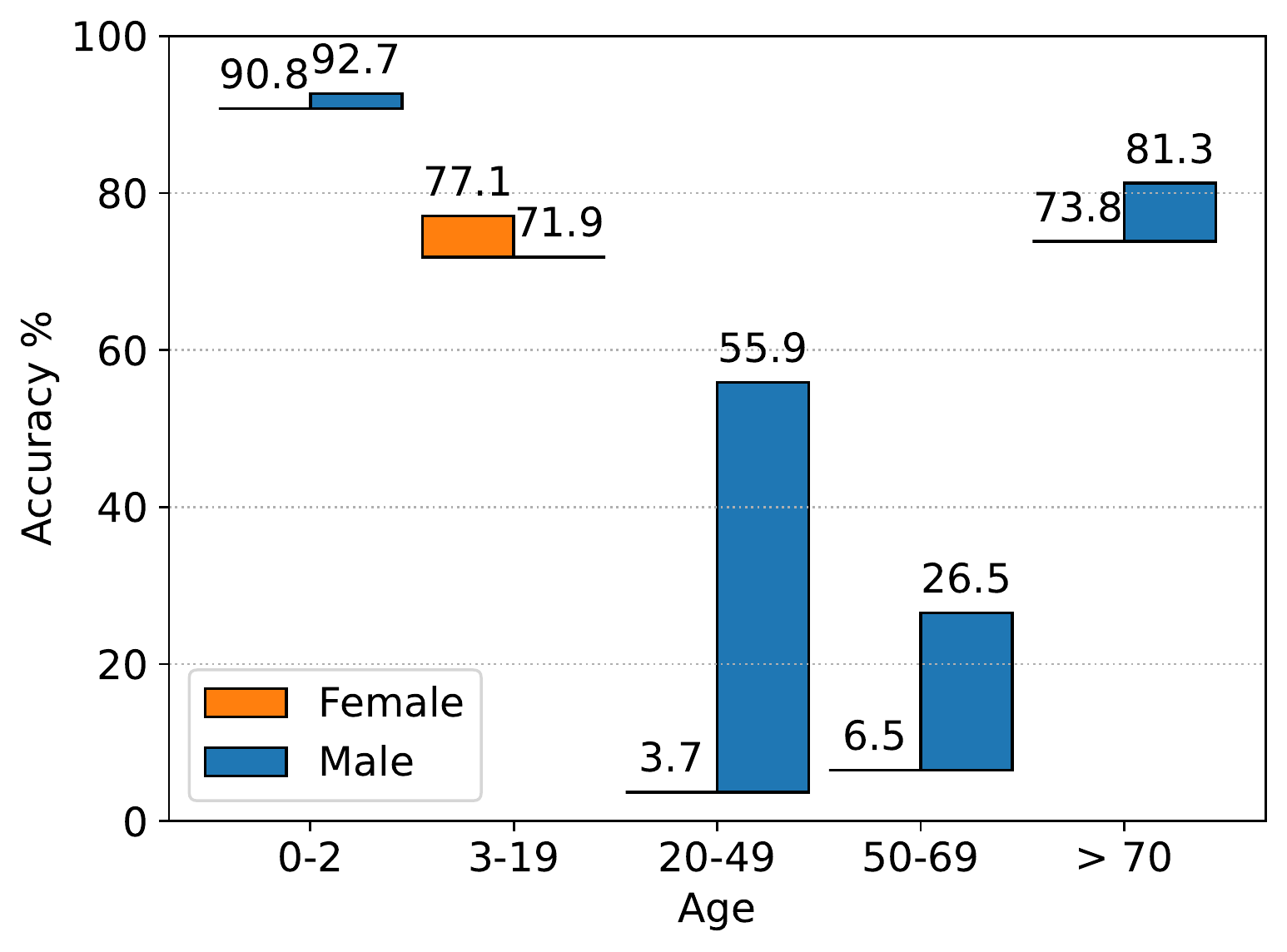}
        \caption{Japanese}
    \end{subfigure}
    \begin{subfigure}[b]{0.24\textwidth}
        \includegraphics[width=\linewidth]{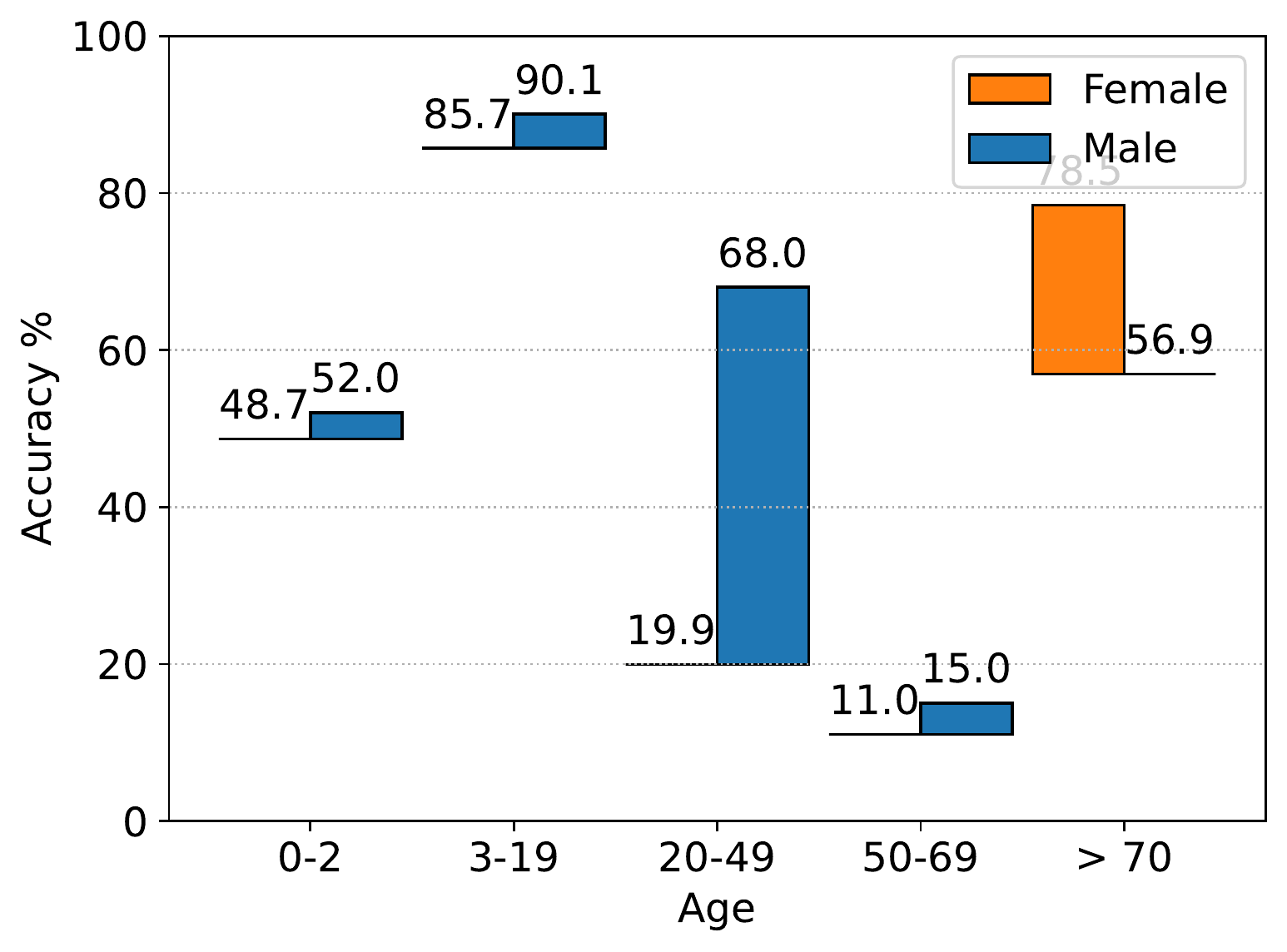}
        \caption{Turkish}
    \end{subfigure}
    \hfill
    \begin{subfigure}[b]{0.24\textwidth}
        \includegraphics[width=\linewidth]{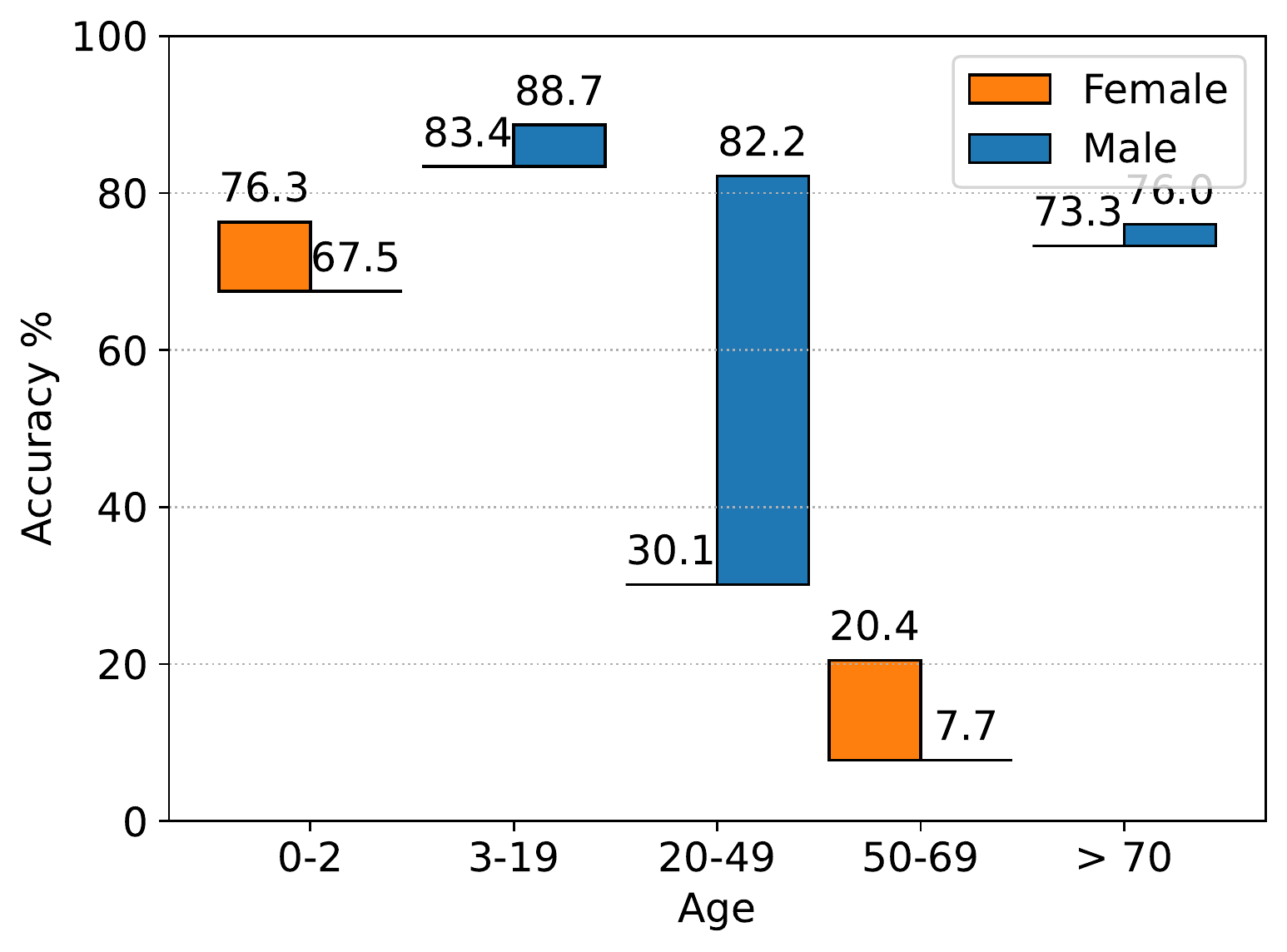}
        \caption{Russian}
    \end{subfigure}
    \hfill
    \begin{subfigure}[b]{0.24\textwidth}
        \includegraphics[width=\linewidth]{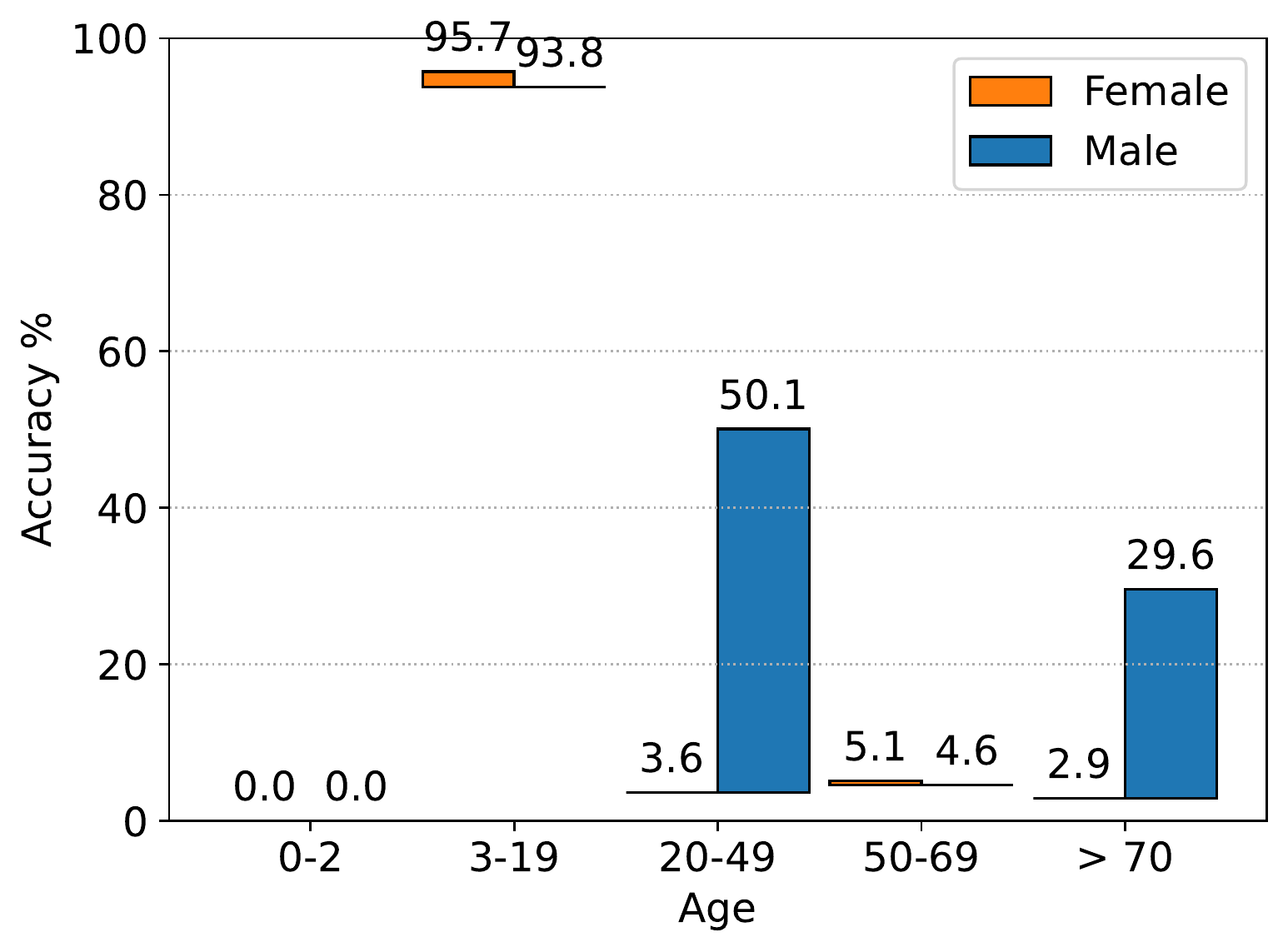}
        \caption{Spanish}
    \end{subfigure}
    \hfill
    \begin{subfigure}[b]{0.24\textwidth}
        \includegraphics[width=\linewidth]{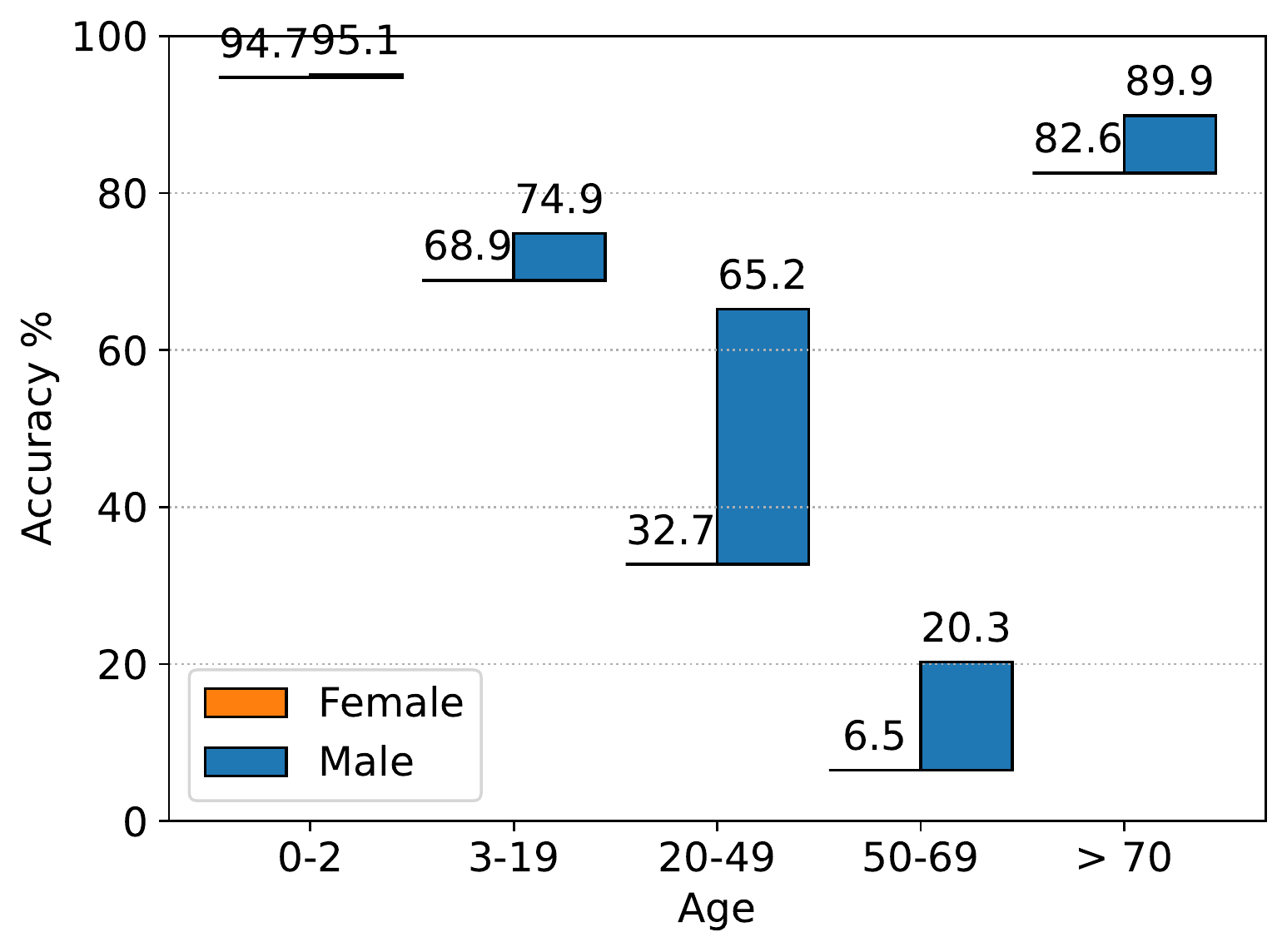}
        \caption{Chinese}
    \end{subfigure}
    \caption{\textbf{Age classification accuracy across female and male groups for different languages.} The blue bars indicate that the male group has higher accuracy than the female group, while orange bars indicate that the female group has higher accuracy. The heights of bars represent the accuracy gaps between male and female groups.}
    \label{fig:age_disparity_across_age}
\end{figure*}

We evaluate multilingual group rate disparity for gender classification on Fairface dataset. We follow the same setup as described in \cref{sec:exp-accuracy-disparity} and measure the gender gap given by \cref{eq:disp-definition}, where $a$ is the composition of male and various race groups, $b$ is the composition of female and various race groups. We defer the complete results to \cref{tab:gender-classification} in \cref{sec:additional-results}. We try to answer the following research questions:
\begin{figure}[!t]
    \centering
    \includegraphics[width=0.9\linewidth]{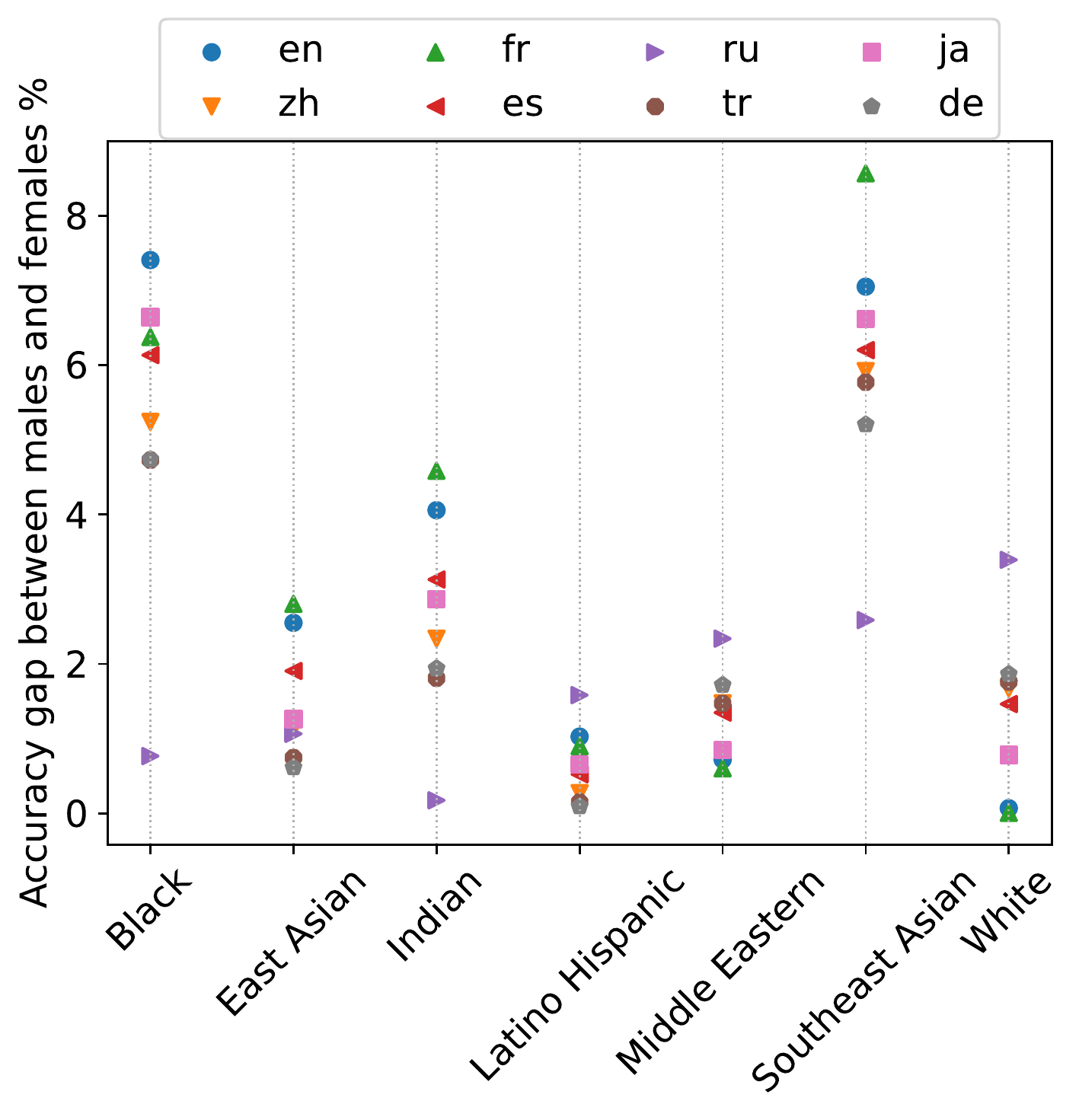}
    \caption{\textbf{Gender accuracy gap across different languages and racial groups.} Black and Southeast Asian people face significant larger gender gaps than other racial groups in most languages.}
    \label{fig:gender_gap_across_race_languages} 
\end{figure}
\begin{table*}[!htb]
\small
\definecolor{goodcolor}{HTML}{BAFFCD}
\definecolor{badcolor}{HTML}{FFC8BA}
\newcommand{\good}[1]{\ccell{goodcolor}{#1}} 
\newcommand{\bad}[1]{\ccell{badcolor}{#1}} 
\newcommand\crule[3][black]{\textcolor{#1}{\rule{#2}{#3}}}
    \centering
    \caption{\textbf{Gender classification accuracy of FairFace images by race groups across different languages.} We note the maximal gender gap across races with \underline{underline} and the maximal gender gap across languages in \textbf{bold}. Taking English as the pivot language, we also highlight any amplified gap compared to English in \bad{red} and any mitigated gap in \good{green}.}
    \label{tab:gender-classification}
    \resizebox{!}{!}{
    \begin{tabular}{l l c c c c c c c c}
    \toprule
        & & & & & East & Southeast & Middle \\
        Language & Gender & White & Black & Indian & Asian & Asian & Eastern & Latino & Average  \\
    \midrule
    \midrule
        English & Female & $95.1$ & $90.9$ & $94.5$ & $95.2$ & $96.0$ & $96.0$ & $94.2$  & 94.6 \\
        & Male & $95.2$ & $83.5$ & $90.4$ & $92.7$ & $89.0$ & $96.7$ & $93.2$  & 91.5 \\
        & $\disp$  & $0.1$ & \textbf{\underline{7.4}} & $4.1$ & $2.5$ & $7.0$ & $0.7$ & $1.0$  & 3.0 \\
    \midrule
        German & Female & $93.8$ & $90.1$ & $94.0$ & $94.2$ & $95.0$ & $95.5$ & $93.9$  & 93.8 \\
        & Male & $95.6$ & $85.4$ & $92.0$ & $93.6$ & $89.8$ & $97.2$ & $93.9$  & 92.5 \\
        & $\disp$  & \bad{$1.9$} & \good{$4.7$} & \good{$1.9$} & \good{$0.6$} & \good{\underline{5.2}} & \bad{$1.7$} & \good{$0.1$}  & \good{$1.3$} \\
    \midrule
        French & Female & $95.0$ & $90.4$ & $94.6$ & $95.0$ & $96.3$ & $95.7$ & $94.2$  & 94.5 \\
        & Male & $95.0$ & $84.0$ & $90.0$ & $92.1$ & $87.8$ & $96.3$ & $93.3$  & 91.2 \\
        & $\disp$  & \good{$0.0$} & \good{\underline{$6.4$}} & \bad{$\mathbf{4.6}$} & \bad{$\mathbf{2.8}$} & \bad{$\mathbf{8.6}$} & \good{$0.6$} & \good{$0.9$}  & \bad{\textbf{3.2}} \\
    \midrule
        Japanese & Female & $94.5$ & $90.6$ & $94.4$ & $94.7$ & $95.7$ & $95.7$ & $94.1$  & 94.2 \\
        & Male & $95.3$ & $84.0$ & $91.5$ & $93.4$ & $89.1$ & $96.6$ & $93.4$  & 91.9 \\
        & $\disp$  & \bad{$0.8$} & \good{\underline{$6.6$}} & \good{$2.9$} & \good{$1.3$} & \good{\underline{$6.6$}} & \bad{$0.8$} & \good{$0.7$}  & \good{$2.3$} \\
    \midrule
        Turkish & Female & $93.9$ & $90.0$ & $93.8$ & $94.6$ & $95.3$ & $95.5$ & $94.1$  & 93.9 \\
        & Male & $95.6$ & $85.2$ & $92.0$ & $93.8$ & $89.5$ & $96.9$ & $93.9$  & 92.4 \\
        & $\disp$  & \bad{$1.8$} & \good{$4.7$} & \good{$1.8$} & \good{$0.7$} & \good{\underline{$5.8$}} & \bad{$1.5$} & \good{$0.1$}  & \good{1.4} \\
    \midrule
        Russian & Female & $93.0$ & $88.4$ & $93.1$ & $93.4$ & $94.6$ & $95.2$ & $93.4$  & 93.0 \\
        & Male & $96.4$ & $87.6$ & $93.2$ & $94.5$ & $92.0$ & $97.5$ & $95.0$  & 93.7 \\
        & $\disp$  & \bad{\textbf{\underline{3.4}}} & \good{$0.8$} & \good{$0.2$} & \good{$1.1$} & \good{$2.6$} & \bad{$\mathbf{2.3}$} & \bad{$\mathbf{1.6}$}  & \good{$0.7$} \\
    \midrule
        Spainish & Female & $94.1$ & $90.5$ & $94.4$ & $95.1$ & $95.6$ & $95.5$ & $94.2$  & 94.2 \\
        & Male & $95.5$ & $84.4$ & $91.2$ & $93.2$ & $89.4$ & $96.8$ & $93.7$  & 92.0 \\
        & $\disp$  & \bad{$1.5$} & \good{$6.1$} & \good{$3.1$} & \good{$1.9$} & \good{\underline{$6.2$}} & \bad{$1.3$} & \good{$0.5$} & \good{$2.2$} \\
    \midrule
        Chinese & Female & $93.9$ & $90.1$ & $94.1$ & $94.8$ & $95.4$ & $95.5$ & $94.2$  & 94.0 \\
        & Male & $95.5$ & $84.9$ & $91.8$ & $93.7$ & $89.5$ & $96.9$ & $93.9$  & 92.3 \\
        & $\disp$  & \bad{$1.7$} & \good{$5.2$} & \good{$2.3$} & \good{$1.1$} & \good{\underline{$5.9$}} & \bad{$1.5$} & \good{$0.3$}  & \good{$1.7$} \\
    \bottomrule
    \end{tabular}
    }
\end{table*}
\squishlist
\item \textbf{How do gender gaps differ across protected groups?} We plot the gender accuracy gap across different languages and racial groups in \cref{fig:gender_gap_across_race_languages}. It is clearly shown that Black and Southeast Asian groups dominantly exhibit larger gender gaps than other groups. We also observe that French has a similar performance with English. We conjecture this is because English and French share the same alphabet and similar syntactic structures. Besides, as shown in \cref{tab:gender-classification}, English and French have the largest race inequality regarding gender gap---nearly zero gender gaps for White but near the maximal gaps for Black.
\item \textbf{Are gender gaps amplified for different languages when compared with English?} We report the accuracy gap on gender classification of FairFace images by race groups across different languages in \cref{tab:gender-classification}. We take English as the pivot language and examine whether the accuracy gaps by race groups are amplified for other languages. Compared with English, accuracy gaps for White and Middle Eastern groups are generally amplified for other languages. On the other hand, accuracy gaps are generally mitigated for groups including Black, Indian, East Asian, Southeast Asian, and Latino groups. The averaged cross-lingual gaps are mitigated for all the languages except for French.
\squishend
We also evaluate multilingual group rate disparity for age classification. We composite gender and age as the group membership. We plot the age classification accuracy by female and male groups across different languages in \cref{fig:age_disparity_across_age}. The blue bars indicate that the male group has higher accuracy than the female group, while the orange bars indicate that the female group has higher accuracy than the male group. The heights of bars represent the accuracy gaps between male and female groups. In general, the male group has higher accuracy than the female group. Especially, adults (20–49 years old) consistently suffer huge gender gaps across all the languages, with the largest gap $52.2\%$ for Japanese.
It is worth noting that the numerals to express ages are identical in text prompts for different languages, e.g., ``a person aged 20 to 49 years'' in English versus ``eine Person im Alter von 20 bis 49 Jahren'' in German. This controlled experiment helps us better understand whether the identical numeric digits have distinct meanings in multilingual contexts. As shown in \cref{fig:age_disparity_across_age}, although text prompts in different languages share the same numerals of ages, the yielding accuracy exhibits significant disparity across languages. One prominent example is that the predictive accuracy for infants (0–2 years old) is $5.8\%$ for English and $2.6\%$ for French, but $89.4\%$ for German and $91.6\%$ for Japanese, implying the presence of significant cross-lingual accuracy gaps.

\section{Conclusion and Limitation}\label{sec:discussion}
Our work extends a growing body of fairness discourse in multilingual and multimodal learning to explore how the multilingual fairness notions, characterized by individual fairness and group fairness, are formulated on the multimodal representations. We stress that multimodal representations are individually fair, but do not prevent accuracy disparity across groups. Our extensive experimental results reveal the negative impacts caused by carelessly applying pre-trained general-purpose multimodal representations. Just one example of this, as discussed in \cref{sec:multilingual-group-rate-disparity-result}, is the significant disparities between cross-lingual gender gaps occurred in age classification. We believe the findings and insights gained through this work can encourage future work to investigate how to mitigate multidimensional biases in representation learning and prevent disparities in the downstream decision-making process.

Our work also has limitations. This work does not provide a thorough explanation on whether the biases and disparities result from the multilingual model itself, or from the datasets it is pre-trained on. However, to give a convincing explanation, it requires either access to large amounts of training data with privacy concerns (the complete datasets for training CLIP are not released yet), or ample computational resources for reproducing the training process. This research question itself is important and worth investigating further.

\section*{Broader Impact}
This work provides insights into fairness in the context of multilingual and multimodal representations. We recognize potential ethical concerns that may arise in the evaluation and address them below.

Firstly, the empirical evaluation for multilingual group fairness adopts the categories of protected groups introduced in the FairFace dataset~\citep{Krkkinen2019FairFaceFA}. We are aware that gender can be non-binary, and individuals can be self-identified outside male and female. Some terms of race attributes in the dataset, such as Latino and Hispanic, are rooted in culture and ethnicity and should not be treated as racial categories. In addition, facial images of low population groups, including Hawaiian and Pacific Islanders and Native Americans, are discarded during data collection. The sensitive attributes in the original FairFace datasets are identified and annotated by human crowd workers. It is possible that the labels of gender, race, and age contain implicit biases and noises. However, these ethical concerns arise from the data collection of the FairFace dataset per se. We anticipate that the methodology we adopted to study multilingual fairness can still be generalized to other data source when more inclusive data collections are available.

Secondly, image classification on the FairFace dataset relies on human-crafted text prompts. The fashion of prompt engineering can be dated from pre-training image and text representations with natural language supervision~\citep{Li2017LearningVN,CLIP}. To avoid offensive and harmful speech towards certain protected groups, we construct the text prompts in a descriptive intent and a neutral tone. 

Finally, the intention for performing classification with sensitive attributes is to validate the presence of biases in pre-trained representations rather than to acquire the personal information of people in the images. Both the evaluated pre-trained models and the benchmark datasets are publicly accessible, and we carefully follow their licenses and agreements for usage. In this sense, we do not foresee any data privacy or information security issues.

\section*{Acknowledgements}
The authors would like to thank Jieyu Zhao, Kai-Wei Chang, Lise Getoor, Jess Law, Hao Tan, and the anonymous reviewers for their fruitful discussions and constructive feedback. This work is supported by the UCSC Applied Artificial Intelligence Initiative (AAII) and the National Science Foundation (NSF) under grant IIS-2040800.

\bibliography{anthology,custom}
\bibliographystyle{acl_natbib}

\clearpage
\appendix
\section{Omitted Proofs}
\subsection{Proof of Lemma~\ref{lem:individual-fairness}}\label{proof:individual-fairness}
\begin{proof}
Given by the definition of cosine similarity, we have
\begin{equation}\label{eq:similarity-gap}
\begin{split}
    & |S(\vb*{v}, \vb*{t}^{(L)}) - S(\vb*{v}, \vb*{t}^{(L')})| \\ 
     =~ & |\frac{\vb*{v}\cdot\vb*{t}^{(L)}}{\|\vb*{v}\|\|\vb*{t}^{(L)}\|} - \frac{\vb*{v}\cdot\vb*{t}^{(L')}}{\|\vb*{v}\|\|\vb*{t}^{(L')}\|}| \\
    =~ &  \frac{|\vb*{v} \cdot (\|\vb*{t}^{(L')}\|\vb*{t}^{(L)} - \|\vb*{t}^{(L)}\|\vb*{t}^{(L')}) |}{\|\vb*{v}\|\|\vb*{t}^{(L)}\|\|\vb*{t}^{(L')}\| }
\end{split}
\end{equation}
From the definition of dot product,
\begin{multline}\label{eq:dot-product}
|\vb*{v} \cdot (\|\vb*{t}^{(L')}\|\vb*{t}^{(L)} - \|\vb*{t}^{(L)}\|\vb*{t}^{(L')})| \leq \\ \|\vb*{v}\| \cdot \|(\|\vb*{t}^{(L')}\|\vb*{t}^{(L)} - \|\vb*{t}^{(L)}\|\vb*{t}^{(L')})\|
\end{multline}
We plug \cref{eq:dot-product} into \cref{eq:similarity-gap} and eliminate the variable $\vb*{v}$
\begin{multline}\label{eq:eliminate-v}
     |S(\vb*{v}, \vb*{t}^{(L)}) - S(\vb*{v}, \vb*{t}^{(L')})| \leq \\ \frac{\|(\|\vb*{t}^{(L')}\|\vb*{t}^{(L)} - \|\vb*{t}^{(L)}\|\vb*{t}^{(L')})\|}{\|\vb*{t}^{(L)}\|\|\vb*{t}^{(L')}\|}
\end{multline}
Let $\theta$ denote the angle between $\vb*{t}^{(L)}$ and $\vb*{t}^{(L')}$, i.e., $$\cos \theta = \frac{\vb*{t}^{(L)} \cdot \vb*{t}^{(L')}}{\|\vb*{t}^{(L)}\|\|\vb*{t}^{(L')}\|},$$ the square of numerator in \cref{eq:eliminate-v} expands as 
\begin{multline}\label{eq:square}
    (\|\vb*{t}^{(L')}\|\vb*{t}^{(L)} - \|\vb*{t}^{(L)}\|\vb*{t}^{(L')})^2 \\ = 2 \|\vb*{t}^{(L)}\|^2 \|\vb*{t}^{(L')}\|^2 ( 1- \cos \theta)
\end{multline}
Substituting the square root of \cref{eq:square} into \cref{eq:eliminate-v}, we eliminate the denominator and obtain
\begin{equation}\label{eq:cosine}
|S(\vb*{v}, \vb*{t}^{(L)}) - S(\vb*{v}, \vb*{t}^{(L')})| \leq \sqrt{2 ( 1 - \cos \theta) }
\end{equation}
Recall that $\vb*{t}^{(L')} \in \mathcal{O}_{\rho}(\vb*{t}^{(L)})$, we can bound $\theta$ by the law of sines
\begin{equation}\label{eq:sine}
    \sup_\theta |\sin \theta| = \sup_{\vb*{t}^{(L')}} \frac{\|\vb*{t}^{(L')} - \vb*{t}^{(L)}\|}{\|\vb*{t}^{(L)}\|} = \frac{\rho}{\|\vb*{t}^{(L)}\|}
\end{equation}
Taking supremums on both sides of \cref{eq:cosine} and combining \cref{eq:sine}, we complete the proof
\begin{align*}
& \sup_{\small \substack{\vb*{t}^{(L')} \in \mathcal{O}_{\rho}(\vb*{t}^{(L)}) \\ 0 \leq \rho < \|\vb*{t}^{(L)}\|}} |S(\vb*{v}, \vb*{t}^{(L')}) - S(\vb*{v}, \vb*{t}^{(L)})| \\
\leq~~ & \sup_{\theta}\sqrt{2 (1 - \sqrt{1 - sin^2 \theta})} \\
=~~ & \sqrt{2 (1 - \sqrt{1 - (\frac{\rho}{\|\vb*{t}^{(L)}\|})^2})}
\end{align*}
\end{proof}

\subsection{Proof of \cref{thm:individual-fairness-approximation}}\label{proof:individual-fairness-approximation}
\begin{proof}
Due to Half-Angle Identities, \cref{eq:cosine} derives as
\begin{equation}\label{eq:half-sine}
    |S(\vb*{v}, \vb*{t}^{(L')}) - S(\vb*{v}, \vb*{t}^{(L)})| \leq 2 |\sin \frac{\theta}{2}|
\end{equation}
For sufficiently small $\theta$, i.e., $\|\vb*{t}^{(L')} - \vb*{t}^{(L)}\| \ll \|\vb*{t}^{(L)}\|$, we take the first-order Taylor approximation 
\begin{equation}\label{eq:first-order-approximation}
    2|\sin\frac{\theta}{2}| \approx |\theta| \approx |\sin \theta| = \frac{\|\vb*{t}^{(L')} - \vb*{t}^{(L)}\|}{\|\vb*{t}^{(L)}\|}
\end{equation}
Combining \cref{eq:half-sine} and \cref{eq:first-order-approximation} we complete the proof.
\end{proof}

\subsection{Proof of \cref{prop: group-fairness}}\label{proof:group-fairness}
\begin{proof}
Expanding $|\acc^{(L)}_{a} - \acc^{(L')}_{b}|$ by triangle inequality we have
\begin{equation}\begin{split}\label{eq:triangle-expanding}
    & \ \ |\acc^{(L)}_{a} - \acc^{(L')}_{b}| \\
    = & \ \ |\acc^{(L)}_{a} - \acc^{(L)}+\acc^{(L)} \\ & \qquad-\acc^{(L')}+\acc^{(L')} - \acc^{(L')}_{b}|\\
    \leq & \ \ |\acc^{(L)}_{a} - \acc^{(L)}|+|\acc^{(L)} -\acc^{(L')}| \\ & \qquad +|\acc^{(L')} - \acc^{(L')}_{b}|
\end{split}\end{equation}

Noticing that $\acc^{(L)} = p_a \cdot \acc^{(L)}_a + p_b \cdot \acc^{(L)}_b$ and $p_a + p_b = 1$, we have
\begin{equation}\label{eq:pb-disp-L}
\begin{split}
    & \ \ |\acc^{(L)}_{a} - \acc^{(L)}| \\
    = & \ \ p_b \cdot |\acc^{(L)}_{a}-\acc^{(L)}_{b}| \\ 
    = & \ \ p_b \cdot \disp^{(L)}(a, b)
\end{split}
\end{equation}
Similarly,
\begin{equation}\label{eq:pa-disp-L}
\begin{split}
    &\ \ |\acc^{(L')} - \acc^{(L')}_b| \\ 
    = &\ \ p_a \cdot |\acc^{(L')}_{a}-\acc^{(L')}_{b}| \\
    = &\ \ p_a \cdot \disp^{(L)}(a, b)
\end{split}
\end{equation}
Substituting \cref{eq:gap-definition}, \cref{eq:pb-disp-L}, and \cref{eq:pa-disp-L} into \cref{eq:triangle-expanding} we complete the proof.
\end{proof}

\section{Additional Experimental Results}\label{sec:additional-results}
\subsection{Empirical Evaluation with Dissimilar Images and Text}
We note that the theoretical analysis posed in Theorem \ref{thm:individual-fairness-approximation} does \emph{not} presume how the images are similar to the text. However, the evaluation in Section \ref{sec:individual-fairness-evaluation} only focuses on similar images and text. To complement for evaluation on dissimilar images and text, we measured to what extent the pre-trained model satisfies multilingual individual fairness for dissimilar images and captions in the Appendix B.1. Specifically, we randomly shuffle the images in the data set such that each image is paired with a random pair of English and German captions. Then we compare the similarity gaps between English and German captions with the images in terms of the encoded textual vector distance between English and German. We observe the same trends for dissimilar images and text: (1) The translation portion generally induces a smaller similarity gap than the independent portion. (2) The CLIP model evaluated on both text corpora has similar empirical $\alpha$ values.  
\begin{figure*}[!htb]
    \centering
    \begin{subfigure}[t]{0.45\linewidth}
        \includegraphics[width=\linewidth]{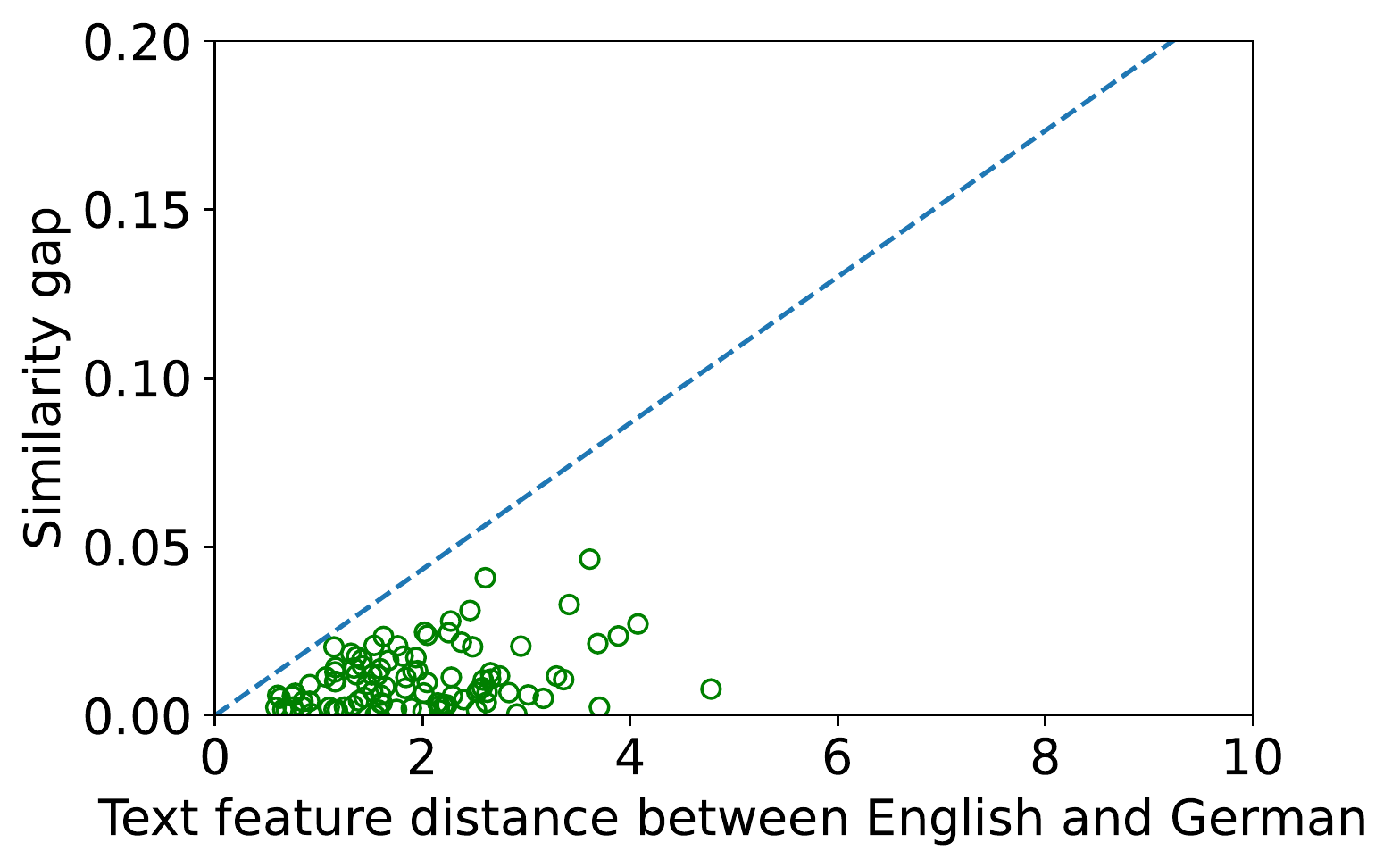}    \caption{Translation}\label{fig:individual-fairness-translation-shuffled}
    \end{subfigure}
    \hfill
    \begin{subfigure}[t]{0.45\linewidth}
        \includegraphics[width=\linewidth]{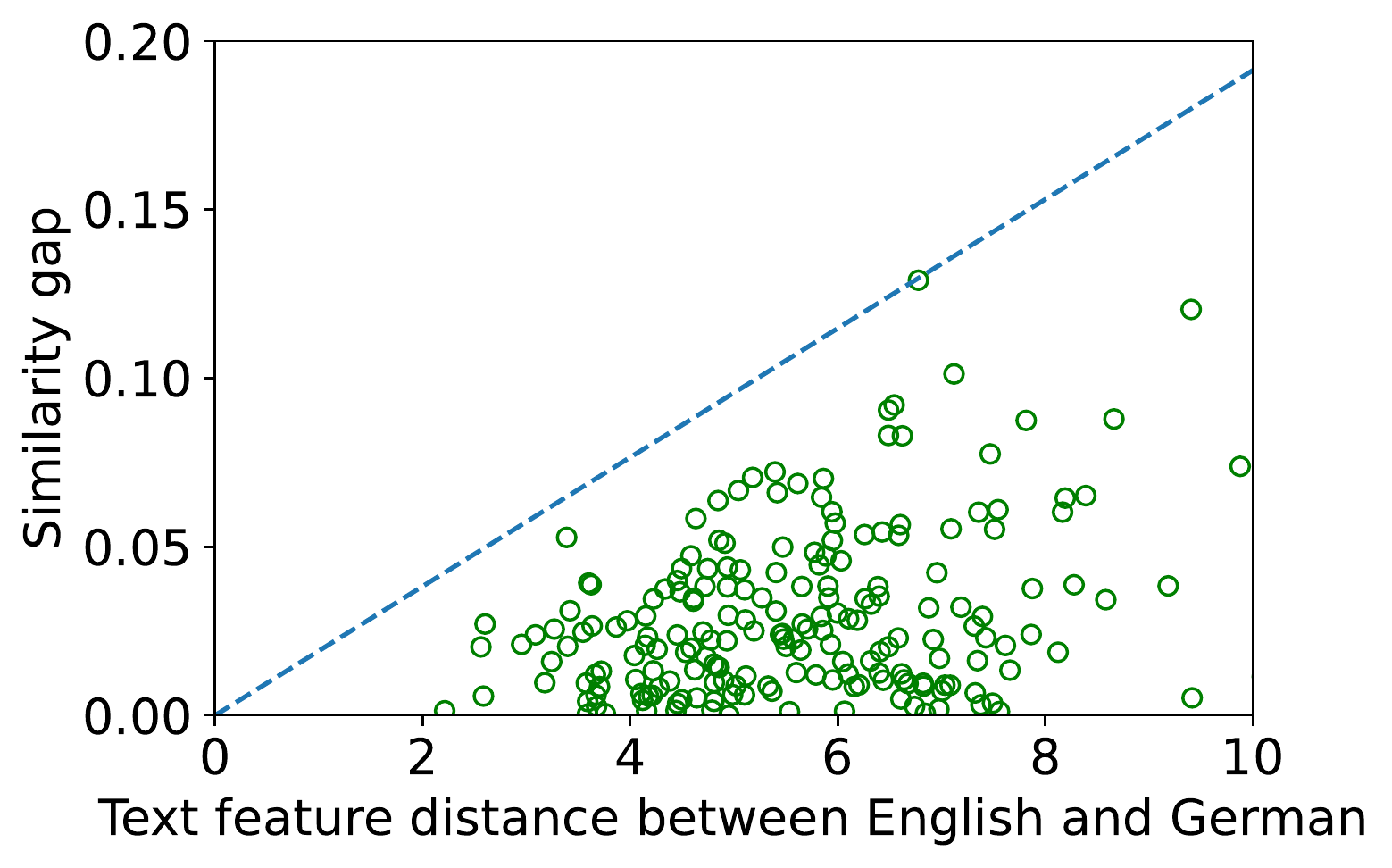}    \caption{Independent}\label{fig:individual-fairness-comparable-shuiffled}
    \end{subfigure}
    \caption{\textbf{We empirically examine how does the multilingual CLIP fare on the translation and the independent portions.} Fig. (\subref{fig:individual-fairness-translation-shuffled}) and (\subref{fig:individual-fairness-comparable-shuiffled}): the $x$-axis represents the distance between English and German captions, the y-axis represents the gap between their corresponding \emph{dissimilarity} scores, and the slope of blue dashed lines represents the empirical $\alpha$ for multilingual individual fairness.
    }
    \label{fig:dissimilar-individual-fairness}
\end{figure*}
\section{Computation Infrastructure}
We use a GPU server with 4 NVIDIA RTX 2080 Ti GPUs for evaluation.

\section{Human Evaluation of the Quality of Machine Translated Text Promts}
\label{sec:AMT}
We recruited crowd workers at Amazon Mechanical Turk (AMT)\footnote{https://www.mturk.com/} to evaluated the quality of text prompts generated in Section \ref{sec:exp-accuracy-disparity}. The crowd workers were supposed to speak both the original language and the translated language to be qualified for completing the tasks. Each task contained one pair of text prompts in the original language (English) and the translated language and was assigned to at least five crowd workers. Each crowd worker was asked to rate the quality of translation from adequacy and fluency on a scale of 1--5. Specifically, we asked the crowd workers the following questions:
\begin{itemize}
    \item \textbf{Adequacy:} does the translated text adequately expresses the meaning in the original text in English?
    \item \textbf{Fluency}: how good the translated language is?
\end{itemize}
\begin{figure*}
    \centering
    \begin{subfigure}{0.45\linewidth}
        \includegraphics[width=\linewidth]{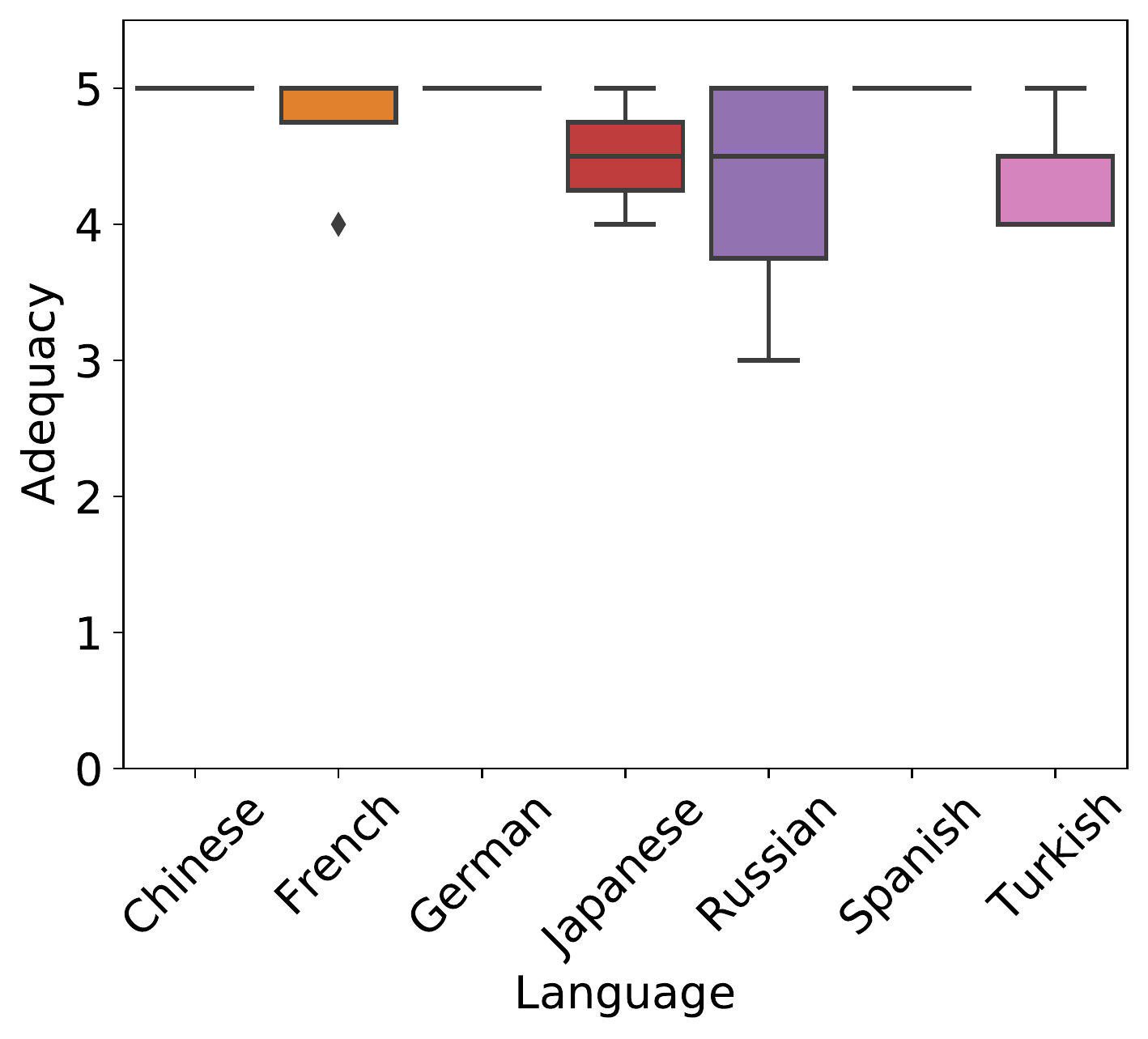}
    \end{subfigure}
    \begin{subfigure}{0.45\linewidth}
        \includegraphics[width=\linewidth]{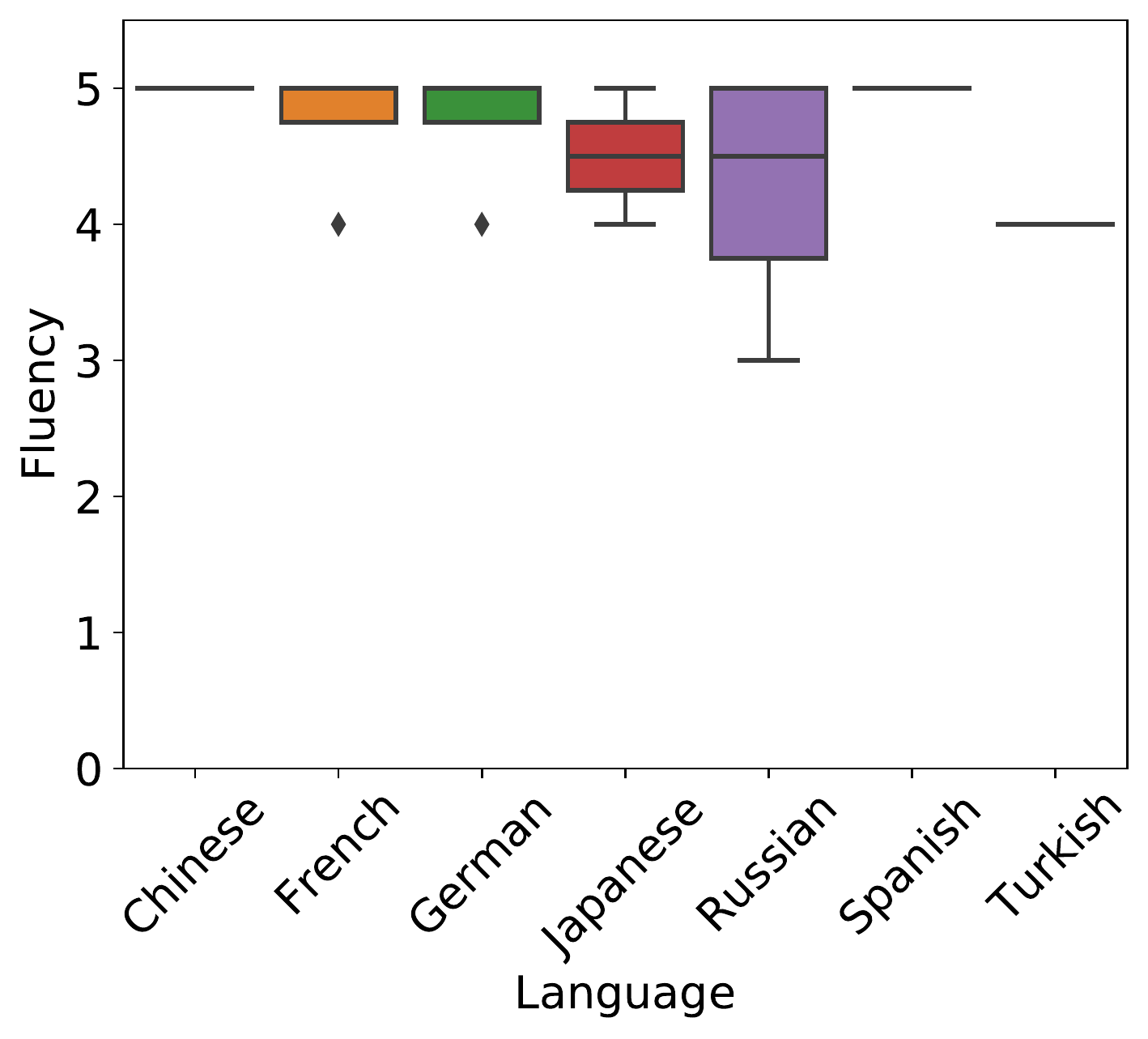}
    \end{subfigure}
    \caption{We recruited crowd workers at AMT to rate the adequacy and fluency of the machine translated text prompts on a scale of 1--5.}
    \label{fig:amt_evaluation}
\end{figure*}
We also asked the workers to point out and fix any potential problems in the prompts.
We collected and visualized the crowdsourced ratings in \cref{fig:amt_evaluation}. For Chinese, French, German, and Japanese, the crowd workers considered the translated text can adequately express all the meanings retained in the English prompts and is flawless. For Japanese, Russian, and Turkish, the crowd workers considered the translations can convey most of the message in the English prompts and are good in fluency.

\end{document}